\documentclass[10pt,twocolumn,letterpaper]{article}
\usepackage[T1]{fontenc}
\usepackage[pagenumbers]{cvpr}
\usepackage{amsmath,amssymb,booktabs,multirow,graphicx}
\usepackage{algorithm}
\usepackage{algpseudocode}
\usepackage{colortbl}
\usepackage{enumitem}
\usepackage{placeins}
\usepackage{tabularx}
\definecolor{cvprblue}{rgb}{0.21,0.49,0.74}
\usepackage[pagebackref,breaklinks,colorlinks,allcolors=cvprblue]{hyperref}
\usepackage[capitalize]{cleveref}
\graphicspath{{figs/}}
\definecolor{bestresultbg}{RGB}{225,237,249}
\definecolor{placeholderink}{RGB}{112,122,136}
\newcommand{\bestresult}[1]{\cellcolor{bestresultbg}\textbf{#1}}
\newcommand{\resulttablestyle}{%
  \footnotesize
  \renewcommand{\arraystretch}{1.16}%
  \setlength{\tabcolsep}{3pt}%
}

\newcommand{\suppfigure}[2]{%
  \IfFileExists{#1}{\includegraphics[width=#2]{#1}}%
  {\includegraphics[width=0.5\textwidth,height=0.22\textheight,keepaspectratio]{example-image-duck}}%
}

\title{WorldInteract: Controllable Streaming Video Generation for Multi-Material Interactions}
\author{%
  \parbox{\dimexpr\textwidth-2\tabcolsep\relax}{\centering\normalsize
    Ahmed Hatem$^{1,*}$\quad Caoliwen Wang$^{1,*,\dagger}$\quad Robert Hoang$^{1,*}$\quad Siyuan Chen$^{1}$\\[3pt]
    Spencer Szabados$^{1}$\quad Zike Wu$^{1}$\quad Xiyang Tan$^{1}$\quad Shenyifan Lu$^{2}$\\[3pt]
    Rafael Bischof$^{3}$\quad Bernd Bickel$^{3}$\quad Renjie Liao$^{1}$\quad Peter Yichen Chen$^{1,\dagger}$\\[7pt]
    $^{1}$University of British Columbia\\[2pt]
    $^{2}$Georgia Institute of Technology\qquad $^{3}$ETH Zurich\\[5pt]
    {\small $^{*}$Equal contribution.\qquad $^{\dagger}$Corresponding authors.}
  }%
}
\hypersetup{
  pdftitle={WorldInteract: Controllable Streaming Video Generation for Multi-Material Interactions},
  pdfauthor={Ahmed Hatem, Caoliwen Wang, Robert Hoang, Siyuan Chen, Spencer Szabados, Zike Wu, Xiyang Tan, Shenyifan Lu, Rafael Bischof, Bernd Bickel, Renjie Liao, Peter Yichen Chen},
  pdfsubject={arXiv preprint}
}

\begin{document}

\raggedbottom
\makeatletter
\g@addto@macro\@maketitle{%
  \begin{center}
    \includegraphics[width=\textwidth]{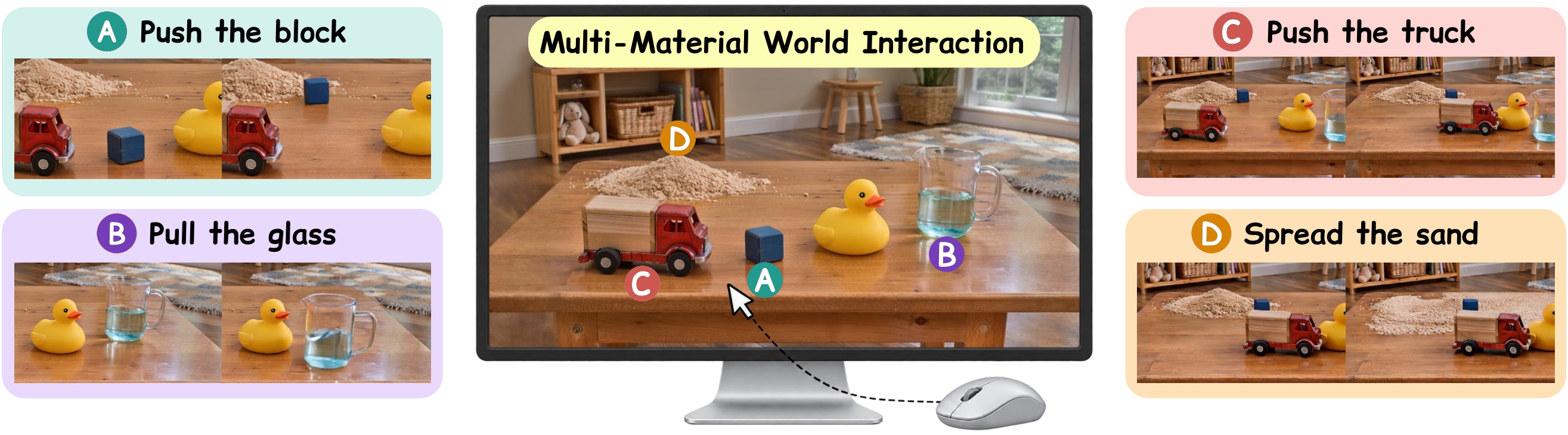}
    \captionsetup{hypcap=false}
    \captionof{figure}{\textbf{WorldInteract} enables interactions among multiple objects with diverse materials during streaming video generation. All four interactions take place within a single video generation process.}
    \captionsetup{hypcap=true}
    \label{fig:teaser}
  \end{center}
}
\makeatother
\maketitle

\begin{abstract}
Video world models aim to create interactive physical worlds that capture how diverse materials move and interact. Yet material-dependent dynamics remain underexplored in controllable video generation, limiting reliable control in scenes with different materials. We present WorldInteract, a streaming video generation framework that decomposes cross-material interaction into material-dependent behavior generation and contact response computation. External pushes and contact impulses are expressed through a shared velocity-increment interface, allowing material behaviors learned from interaction-free videos to be reused in interacting scenes. A physics controller predicts contacts and computes velocity increments from estimated masses and restitution coefficients, while a causal generator executes these increments using learned material behaviors. This separation enables cross-material interaction without supervised contact videos spanning diverse material pairings and contact configurations. We train the generator on 30k simulated interaction-free videos spanning 22 material classes and introduce Material-Aware RL, using verifiable physical rewards to refine material behavior and control execution. WorldInteract substantially outperforms state-of-the-art methods on Physics-IQ and delivers more reliable push control on real-world images.
\end{abstract}

\section{Introduction}
\label{sec:intro}
Interactive video world models aim to respond to user actions with
physically plausible outcomes, yet the same push can roll a ball, slide a
box, or scatter a pile of sand. Mass and friction affect how far an object travels after a push; mass and restitution govern how contact
changes the motion of both objects. Reliable control therefore requires modeling both \emph{material behavior}, which describes how an object moves under an applied velocity change, and \emph{contact response}, which determines how momentum is exchanged when objects collide. Their combination governs cross-material interactions. Ignoring either component makes control unreliable when materials or contact partners change.

Recent work establishes important foundations for physical control.
Force Prompting~\cite{gillman2025forceprompting} and
PhysCtrl~\cite{wang2025physctrl} learn force control from simulated data,
while Goal Force~\cite{goalforce} extends video generation to goals
involving object interactions. StreamForce~\cite{wang2026streamforce} and
PhysStream~\cite{chen2026physstream} support updated controls during
streaming generation. However, systematic modeling of material behavior
and cross-material interaction remains underexplored. Material properties
determine how much motion occurs, how long it persists, and how it
propagates through contact. These dependencies must be captured for
controls to remain physically consistent across different materials.

Learning these behaviors directly from contact videos creates a coverage
problem. Adding a material introduces new pairings with existing materials,
even if its individual behavior is already represented. Each pairing can
also appear under different contact configurations. Supervision must cover
these combinations as well as individual material behaviors, making it
costly to extend control to new materials.

We present \textbf{WorldInteract}, a streaming video generation framework that separates material behavior learning from contact response computation. Our key observation is that
external pushes and contact impulses both change object velocity. A shared
\emph{velocity-increment interface} therefore allows material behaviors
learned from interaction-free videos to be reused in interacting scenes.
At inference, a \emph{physics controller} predicts contacts and computes
velocity increments from estimated masses and restitution coefficients,
while a causal generator executes these increments using learned material behaviors. The controller determines the momentum exchange; the generator
produces the resulting behavior. This decomposition enables interaction
without supervised contact videos spanning material pairings and contact
configurations.

We construct 30k simulated \emph{interaction-free videos}, which contain no
object--object contact, spanning 22 material classes. Scene memory maintains
spatial state and object identity as generation proceeds under updated
controls. Supervised video reconstruction alone does not directly enforce
relative material behaviors or the timing of control events. We introduce
\emph{Material-Aware RL} to refine material behavior and control execution
through verifiable physical rewards on generated rollouts. These rollouts
include contact scenes produced with the physics controller, extending
training beyond the interaction-free supervision. Experiments demonstrate clear
advantages over state-of-the-art methods, with stronger physical consistency
on Physics-IQ and higher push success on real-world images.

Our contributions are:
\begin{itemize}
  \item \textbf{A compositional framework for cross-material interaction.}
  We couple a streaming video generator for material-dependent behavior with
  a physics controller for contact response computation. Their shared
  velocity-increment interface enables cross-material interaction to be
  composed at inference.
  \item \textbf{Material-Aware RL.} We improve velocity-increment execution
  with verifiable rewards that assess material behavior and cross-material
  interaction in generated rollouts.
  \item \textbf{An interaction-free multi-material dataset.} We provide 30k
  simulated videos across 22 material classes to learn material behavior
  from applied pushes, avoiding exhaustive supervision over material
  pairings and contact configurations.
\end{itemize}

\section{Related Work}
\label{sec:related}

\paragraph{Controllable video generation.}
Motion-conditioned video models guide objects through spatial controls.
DragAnything~\cite{wu2024draganything}, Tora~\cite{zhang2024tora}, and
FlashMotion~\cite{li2026flashmotion} generate motion along specified
trajectories, while DragStream~\cite{zhou2025dragstream} and
MotionStream~\cite{shin2025motionstream} allow motion guidance to change
during streaming generation. Such controls specify desired motion, but do
not by themselves determine how material properties and contact alter that
motion. WorldInteract uses velocity increments as physical inputs, leaving
material behavior to the generator and contact response computation to the
physics controller.

\paragraph{Physics-grounded video generation.}
Simulation provides explicit physical structure for visual synthesis.
PhysGen~\cite{liu2024physgen}, PhysGen3D~\cite{chen2025physgen3d},
PhysMotion~\cite{tan2024physmotion}, and PSIVG~\cite{foo2026psivg} use
simulated dynamics to guide generation. RealWonder~\cite{liu2026realwonder}
combines single-image 3D reconstruction, physics simulation, and a distilled
generator for real-time physical control. Another line learns physical
control from simulated videos: Force Prompting~\cite{gillman2025forceprompting}
and PhysCtrl~\cite{wang2025physctrl} learn force-conditioned motion, while
Goal Force~\cite{goalforce} targets physics-conditioned goals involving
object interactions. StreamForce~\cite{wang2026streamforce} supports updated
forces during streaming generation, and PhysStream~\cite{chen2026physstream}
introduces structured scene memory for velocity-increment control.
WorldInteract differs by separating material-dependent behavior generation
from contact response computation. Learning the former from interaction-free videos and computing
the latter online reduces the need for supervision over material pairings
and contact configurations.

\paragraph{Streaming video generation.}
Autoregressive video models support streaming synthesis through incremental
generation and history
conditioning~\cite{kim2024fifodiffusion,henschel2024streamingt2v,chen2024diffusionforcing,song2025historyguided}.
Causal distillation and training on generated histories improve inference
speed and rollout
stability~\cite{yin2024causvid,huang2025selfforcing,causalforcing}.
Further work addresses long-horizon consistency and context
management~\cite{liu2025rollingforcing,yang2025longlive,li2026rollingsink}.
WorldInteract focuses on material-dependent behavior and cross-material
interaction, using velocity increments to connect a physics controller
with the streaming generator.

\section{Method}
\label{sec:method}
\label{sec:method:training}

\begin{figure*}[t]
  \centering
  \ifdefined\XeTeXpdffile
    \XeTeXpdffile "figs/fig_pipeline.pdf" width \textwidth
  \else
    \ifdefined\pdfximage
      \pdfximage width \textwidth {figs/fig_pipeline.pdf}%
      \pdfrefximage\pdflastximage
    \else
      \includegraphics[width=\textwidth]{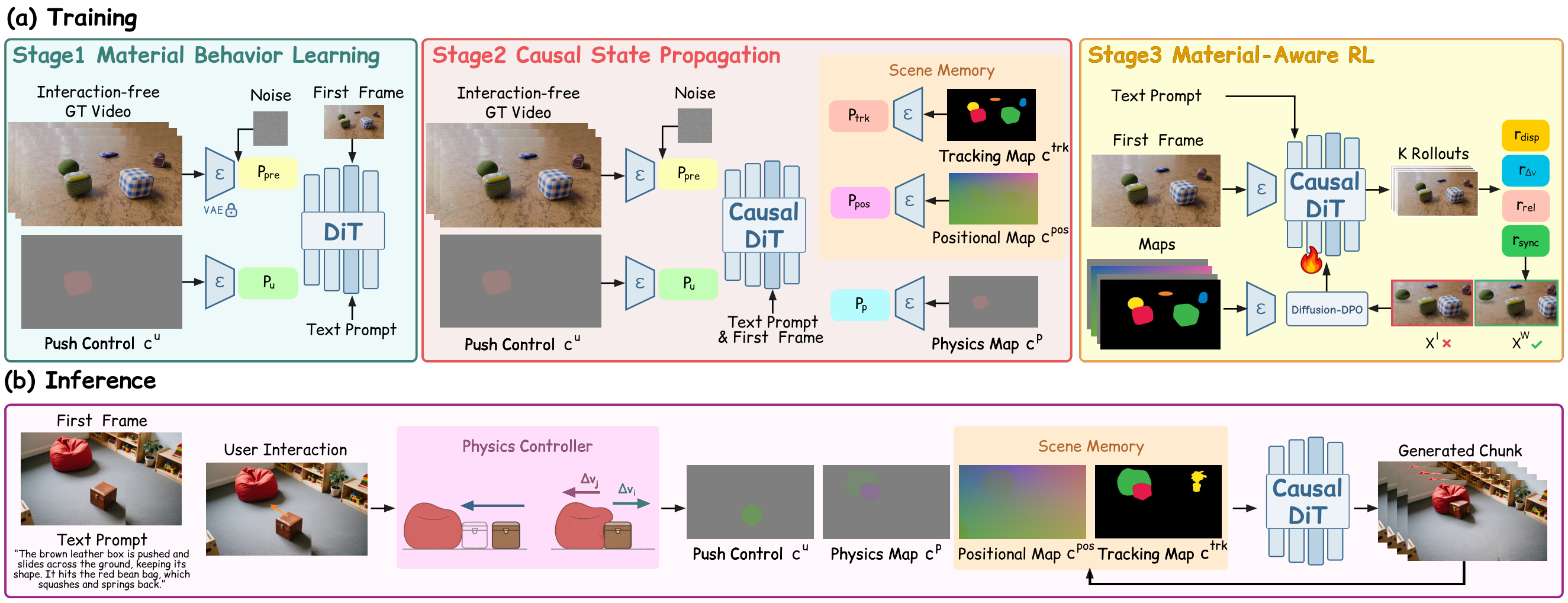}
    \fi
  \fi
  \par
  \captionsetup{position=bottom,skip=8pt}
  \caption{\textbf{WorldInteract pipeline.} (a)~Training. Stage~1 learns material behavior from interaction-free videos with full attention (\cref{sec:method:response}).
Stage~2 adapts the generator to causal generation with scene memory and a
physics map (\cref{sec:method:generator}). Stage~3 refines the generator on its
own rollouts with Material-Aware RL (\cref{sec:method:alignment}).
(b)~Inference (\cref{sec:method:inference}). The physics controller predicts
contacts from previously generated frames and computes velocity increments
$\Delta\mathbf v_i,\Delta\mathbf v_j$, which the causal generator executes
through the physics map $c^{\mathrm p}$. $\mathcal E$ denotes the
VAE encoder.}
  \label{fig:pipeline}
\end{figure*}

As shown in \cref{fig:pipeline}, WorldInteract assigns
material-dependent behavior generation to a video generator and contact
response computation to a physics controller. The controller expresses external pushes and
predicted contact impulses through a shared \emph{velocity-increment
interface}. At inference, it computes these increments from object properties estimated
by a vision-language model (VLM), while the generator, trained on how materials move after a velocity change, executes them. This division allows supervised training to
cover individual material behaviors without enumerating material pairings
and contact configurations.

\paragraph{Training overview.}
Training has three stages. Stage~1 learns material behavior from
interaction-free videos (\cref{sec:method:response}). Stage~2 introduces causal
generation with scene memory and a physics map
(\cref{sec:method:generator}). Stage~3, Material-Aware RL, improves control
execution on generated rollouts (\cref{sec:method:alignment}). The first
two stages use supervised flow matching; the third uses reward-based
preference optimization with supervised replay. At inference, the
controller computes contact responses and the generator executes them
(\cref{sec:method:inference}). Full training objectives are provided in
the Appendix.

\subsection{Material Behavior Learning}
\label{sec:method:response}

This stage learns material-dependent behavior conditioned on user
interactions. We first convert each interaction into a velocity increment,
providing a shared interface for external pushes and contact impulses.
The generator then learns from interaction-free videos how different
materials behave under these increments.

\paragraph{Interaction-free data.}
\label{sec:method:data}

\begin{figure}[t]
  \centering
  \includegraphics[width=\linewidth]{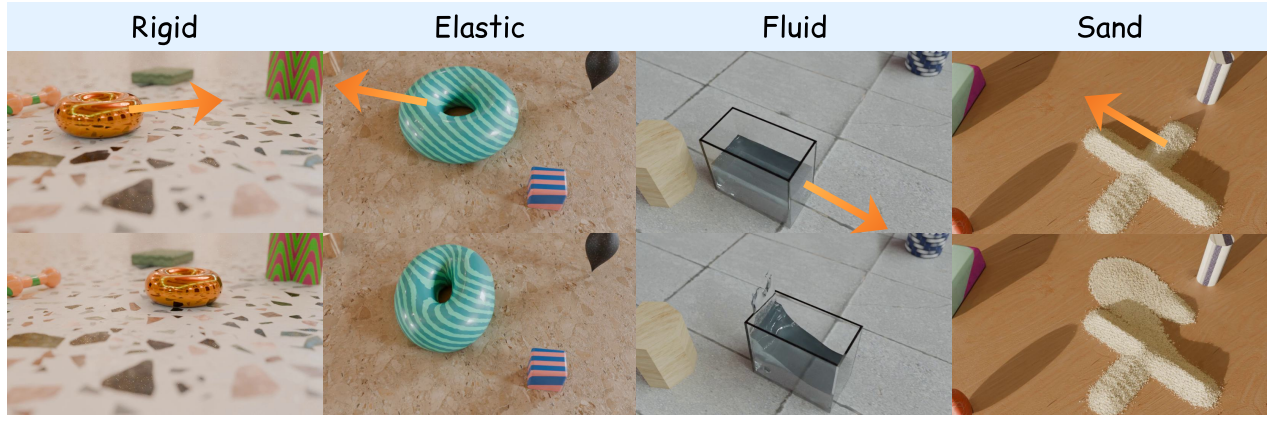}
  \caption{\textbf{Multi-material dataset.} Our dataset contains 30k interaction-free clips spanning 22 material classes, diverse object assets, and randomized environments.}
  \label{fig:dataset}
\end{figure}
We construct 30k simulated \emph{interaction-free videos} spanning diverse
material classes. Figure~\ref{fig:dataset} shows examples of rigid and
elastic solids, fluids, and sand. Here, interaction-free excludes object--object
contact; contact with the ground, friction, and rebounds remain in the
videos.

\paragraph{Velocity-increment interface.}
Let $(t,o,\mathbf J)$ denote a user interaction, where $t$ is the frame
at which a push is applied, $o$ identifies the target object, and
$\mathbf J\in\mathbb R^3$ is the applied impulse. We define the resulting
velocity increment as $\Delta\mathbf v=\mathbf J/m_o$, where $m_o$ is the
object's mass. Training uses the simulator's true mass; at inference, the
controller uses the VLM estimate $\hat m_o$ (\cref{sec:method:inference}).
The shared velocity-increment interface represents both external pushes
and contact impulses, allowing material behaviors learned from pushes to
be reused after contact.

\paragraph{Behavior supervision.}
We encode user interactions as the push control $c^{\mathrm u}$, whose
three channels represent the $x$-, $y$-, and $z$-components of
$\Delta\mathbf v$. Encoding details are provided in the Appendix.

Starting from Wan2.2-TI2V-5B~\cite{wan22}, we add a zero-initialized patch
embedding $P_{\mathrm u}$ for the push control and fine-tune the diffusion
transformer (DiT) on 49-frame clips with full attention. Let $\mathcal E$
denote the VAE encoder, $z_0$ the clean first-frame latent, $z$ the remaining
clean video latents, and $y$ the backbone's text conditioning. We sample
$\tau\sim\mathcal U(0,1)$ and Gaussian noise
$\epsilon\sim\mathcal N(0,I)$, forming
$z^\tau=(1-\tau)z+\tau\epsilon$. We fuse the noisy video and push control
into input tokens:
\begin{equation}
h^\tau=P_{\mathrm{pre}}(z^\tau)
       +P_{\mathrm u}\big(\mathcal E(c^{\mathrm u})\big),
\label{eq:response-fusion}
\end{equation}
where $P_{\mathrm{pre}}$ is the pretrained video patch embedding. The DiT
takes $h^\tau$, conditioned on $\tau$, $z_0$, and $y$, and predicts the
latent-space vector field $\hat v_\theta^\tau(h^\tau,\tau,z_0,y)$ used to
denoise the video. We train the DiT, $P_{\mathrm u}$, $P_{\mathrm{pre}}$,
and the output head with the rectified-flow matching
loss~\cite{flowmatching}:
\begin{equation}
\mathcal F_\theta(z)=\mathbb E_{\tau,\epsilon}
\big\|\hat v_\theta^\tau(h^\tau,\tau,z_0,y)-(\epsilon-z)\big\|_2^2.
\label{eq:fm}
\end{equation}
The VAE, the text encoder, and the text cross-attention keys and values stay frozen.
\subsection{Causal State Propagation}
\label{sec:method:generator}

This stage learns to preserve material-dependent behavior while accepting
new interactions during streaming generation. Scene memory records the
evolving scene, while a physics map specifies velocity increments at
recent object locations.

\paragraph{Causal generator.}
Starting from the Stage-1 checkpoint, we replace full attention with
block-causal attention~\cite{causalforcing}. The causal VAE represents the
first frame as $z_0$ and each subsequent four-frame chunk as $z_\ell$.
The DiT denoises one chunk at a time, attending to earlier latent chunks
through a key--value cache.

\paragraph{Scene memory.}
We represent scene state with two maps: $c^{\mathrm{pos}}$ records 3D
positions derived from Depth-Anything-3~\cite{da3}, and
$c^{\mathrm{trk}}$ assigns each object a persistent identity using
SAM2~\cite{sam2}. Both maps are updated from each decoded chunk, providing
the generator with object locations and identities for the next chunk.

\paragraph{Physics map.}
The three-channel map $c^{\mathrm p}$ encodes velocity increments on
objects' recent masks, allowing control to follow objects as they move.
At inference, it carries velocity increments computed by the physics
controller for predicted contacts. To train this interface without contact videos, we route a
subset of mid-video pushes, including all re-pushes, through
$c^{\mathrm p}$. The branch thus learns to execute the same velocity
increments used for contact response.

\paragraph{Causal supervision.}
As shown in \cref{fig:pipeline}, we VAE-encode the condition maps
and add their patch embeddings to the noisy video tokens:
\begin{equation}
h_\ell^\tau=P_{\mathrm{pre}}(z_\ell^\tau)
 +\textstyle\sum_{m\in\mathcal M}P_m\big(\mathcal E(c^m_{\ell-1})\big),
\label{eq:fusion}
\end{equation}
where $\mathcal M=\{\mathrm u,\mathrm{pos},\mathrm{trk},\mathrm p\}$.
The one-chunk shift ensures that spatial conditions use only earlier
frames. The DiT takes $h_\ell^\tau$, $\tau$, $y$, and preceding latents
to predict the vector field for denoising chunk $\ell$. We train with
teacher forcing, using clean ground-truth history and the flow-matching
loss in \cref{eq:fm}. We train the DiT and the three new,
zero-initialized branches $P_{\mathrm{pos}},P_{\mathrm{trk}},P_{\mathrm p}$, while the Stage-1 layers $P_{\mathrm u}$, $P_{\mathrm{pre}}$, and the output head stay frozen.

\subsection{Material-Aware RL}
\label{sec:method:alignment}

Flow matching does not directly enforce physical relations such as
stronger pushes producing greater displacement. Material-Aware RL uses
verifiable rewards on generated rollouts to refine material behavior
and velocity-increment execution.

\paragraph{Training inputs.}
We use simulated pairs that vary only in material or push magnitude,
real-world image pairs with different push magnitudes, and simulated
initial scenes with objects on a collision course. For each input, we
sample $K=4$ rollouts. In contact scenes, the controller supplies velocity
increments for predicted contacts, providing interaction experience without
supervised contact videos.

\paragraph{Physical rewards.}
SAM2 masks, CoTracker3~\cite{cotracker3} tracks, and depth maps provide
four checks: displacement accuracy ($r_{\mathrm{disp}}$), agreement with
requested velocity increments ($r_{\Delta v}$), relative behavior across
paired inputs ($r_{\mathrm{rel}}$), and event timing ($r_{\mathrm{sync}}$).
Simulated interaction-free inputs provide target displacements. For real-world
images and contact scenes, displacement is scored only for objects
receiving no increment, rewarding unaffected neighbors that stay still.
Contact execution is evaluated against the controller's increments,
so the rewards train the generator to follow a contact response while
the controller remains responsible for computing it.

\paragraph{Preference optimization.}
We select the highest- and lowest-reward rollouts, $x^w$ and $x^l$, among candidates with the same predicted contacts. Diffusion-DPO~\cite{diffusiondpo,dpo}, adapted
to flow matching, increases the preferred rollout's relative likelihood
using a frozen copy of the final Stage-2 model as reference. We train rank-64 LoRA adapters on the DiT's linear layers and fully fine-tune the four condition branches. We also replay flow-matching supervision on simulated videos to limit visual drift.

\paragraph{Acceleration.}
We distill the Stage-3 model into a four-step causal student with
DMD~\cite{dmd,yin2024causvid}. Initializing from the causal teacher avoids
the ODE-regression stage needed when distilling a bidirectional
teacher~\cite{causalforcing}. The distilled model supports interactive streaming generation at 8 fps on a single H100. VLM queries are cached per object (\cref{sec:method:inference}) and do not affect per-chunk speed.

\subsection{Inference}
\label{sec:method:inference}
\label{sec:method:vlm}

As shown in \cref{fig:pipeline}, inference alternates between
the physics controller, which estimates scene state and computes contact
responses, and the video generator, which produces the next chunk.
All state estimates use the initial image or previously generated frames.

\paragraph{Physics controller.}
For each object $k$, the controller queries the VLM to estimate mass
$\hat m_k$ and restitution coefficient $\hat e_k$ from its image crop and
approximate metric size. These properties are queried when the object enters a contact and cached for later interactions.

Depth and object tracks provide each object's 3D position, velocity
$\mathbf v_k$, and acceleration. After incorporating user pushes, the
controller extrapolates these states over the next chunk and predicts
contacts where projected masks and depth intervals overlap. Only
approaching contacts ($v_n>0$, defined below) trigger impulses; resting
contacts are left to the generator's learned behavior.

For contacting objects $i$ and $j$, let $\mathbf n$ point from $j$ to $i$
along their centroid line, and let
$v_n=(\mathbf v_j-\mathbf v_i)\cdot\mathbf n$ be their closing speed.
With combined restitution $\hat e=\hat e_i\hat e_j$, a frictionless central collision
model~\cite{pybullet} gives
\begin{equation}
\begin{aligned}
\Delta\mathbf v_i&=\phantom{-}(1+\hat e)\tfrac{\hat m_j}{\hat m_i+\hat m_j}\,v_n\mathbf n,\\
\Delta\mathbf v_j&=-(1+\hat e)\tfrac{\hat m_i}{\hat m_i+\hat m_j}\,v_n\mathbf n.
\end{aligned}
\label{eq:impulse}
\end{equation}
After each predicted contact, the controller updates both objects'
extrapolated velocities and re-predicts later contacts, propagating
velocity changes along contact chains. It writes the contact increments
to $c^{\mathrm p}$ using recent masks and the one-chunk shift defined
above. Contact thresholds, map timing, and simultaneous-contact handling
are specified in the Appendix.

\paragraph{Video generator.}
The DiT denoises the next chunk
using the user push control, updated scene memory, and physics map.
The decoded frames become the history for the next update.

\section{Experiments}
\label{sec:experiments}

\begin{figure*}[t]
  \centering
  \includegraphics[width=\textwidth]{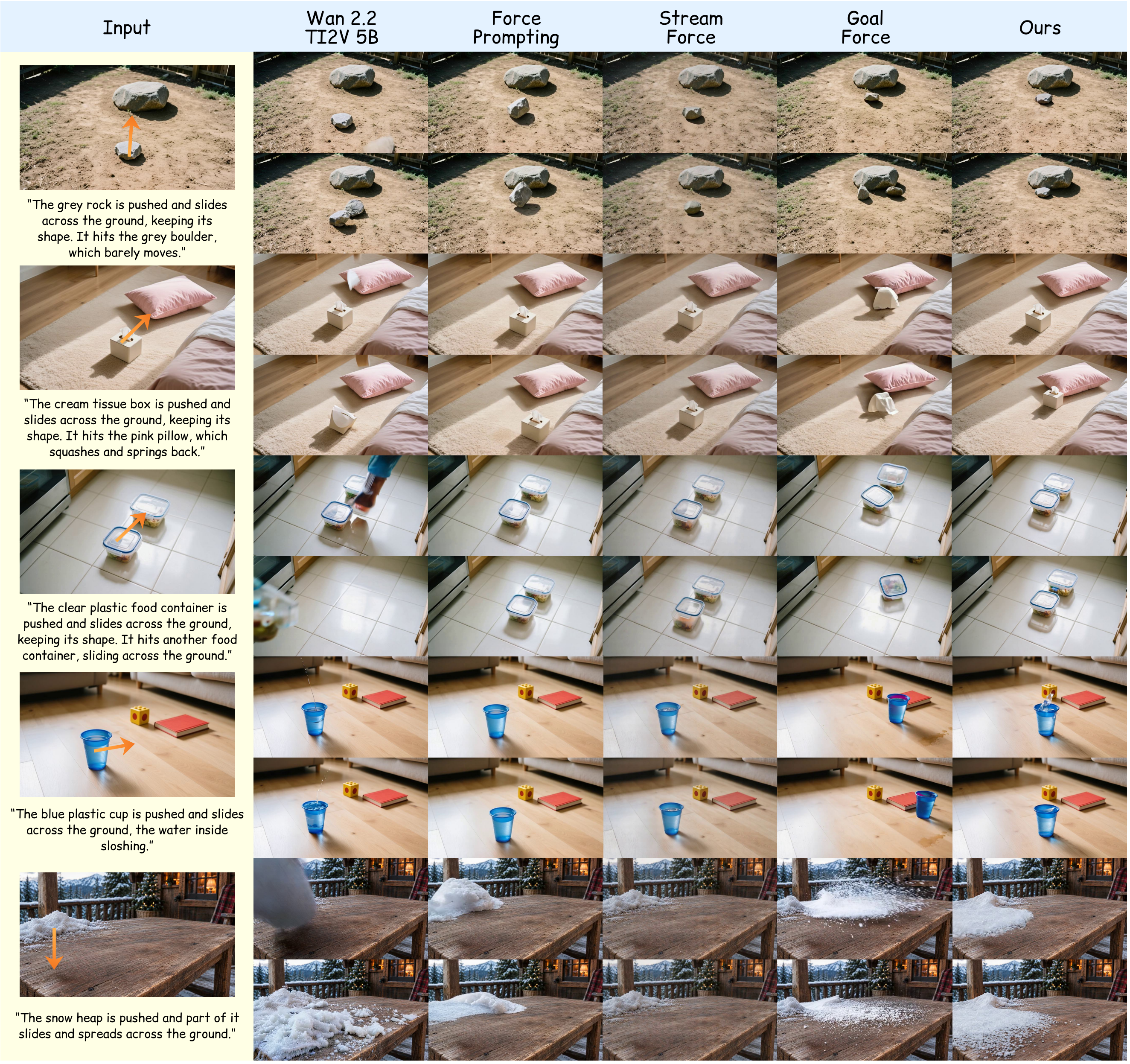}
  \caption{\textbf{Qualitative comparison with baselines.} Top and bottom
rows show the middle and final frames of each scene. Scenes span rigid,
deformable, fluid, and granular materials under the push shown in the Input column.}
  \label{fig:qualitative_nonstream}
\end{figure*}

\begin{figure*}[t]
  \centering
\includegraphics[width=0.9\textwidth]{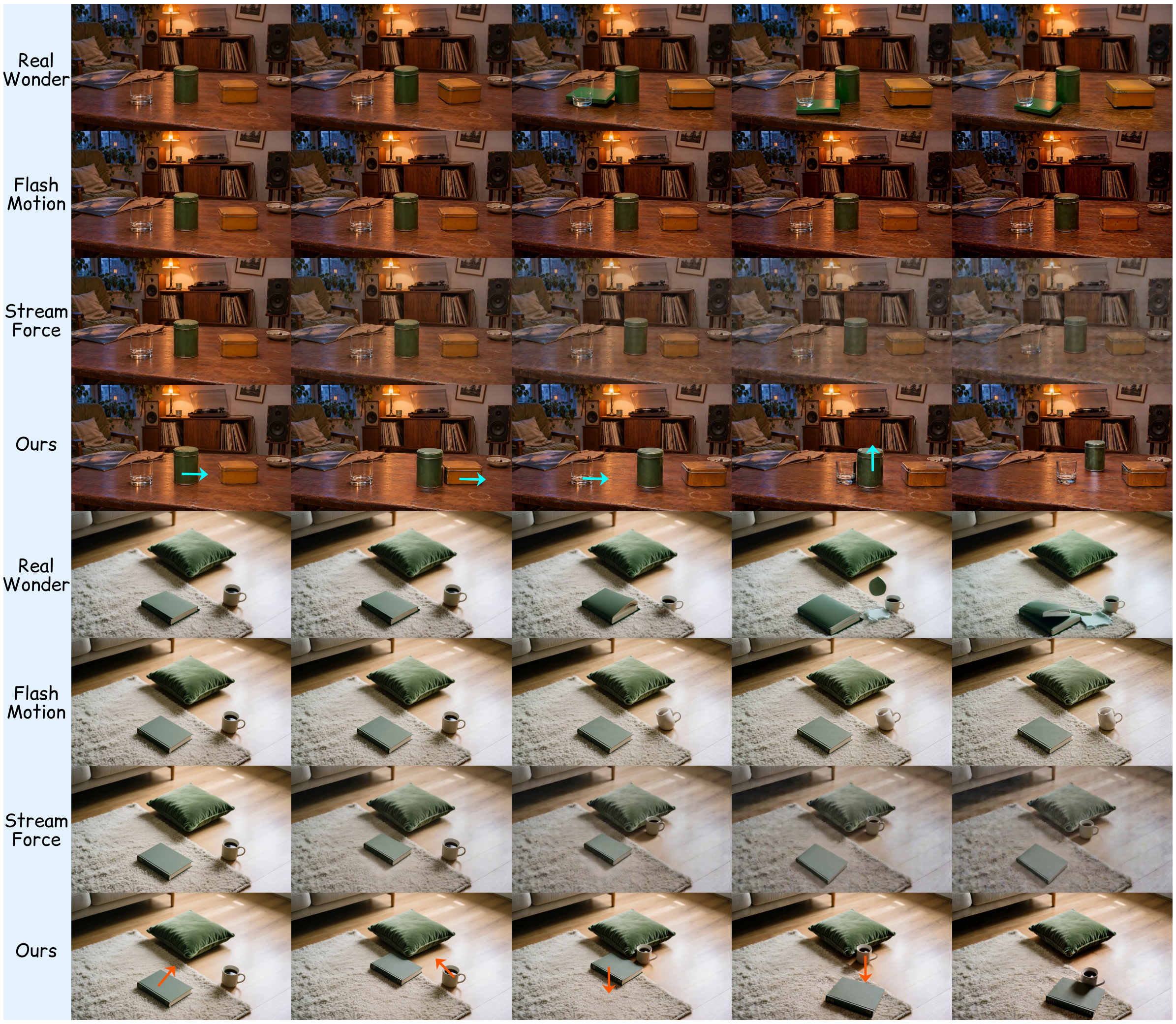}
  \caption{\textbf{Qualitative comparison with streaming baselines.} Frames
progress left to right within a single rollout, and arrows mark pushes
issued at different times. Each push targets an object that has already
moved. Baselines lose track of the scene over time, while WorldInteract
executes each push at the object's current position.}
  \label{fig:qualitative_stream}
\end{figure*}

\begin{figure*}[t]
  \centering
  \includegraphics[width=0.95\textwidth]{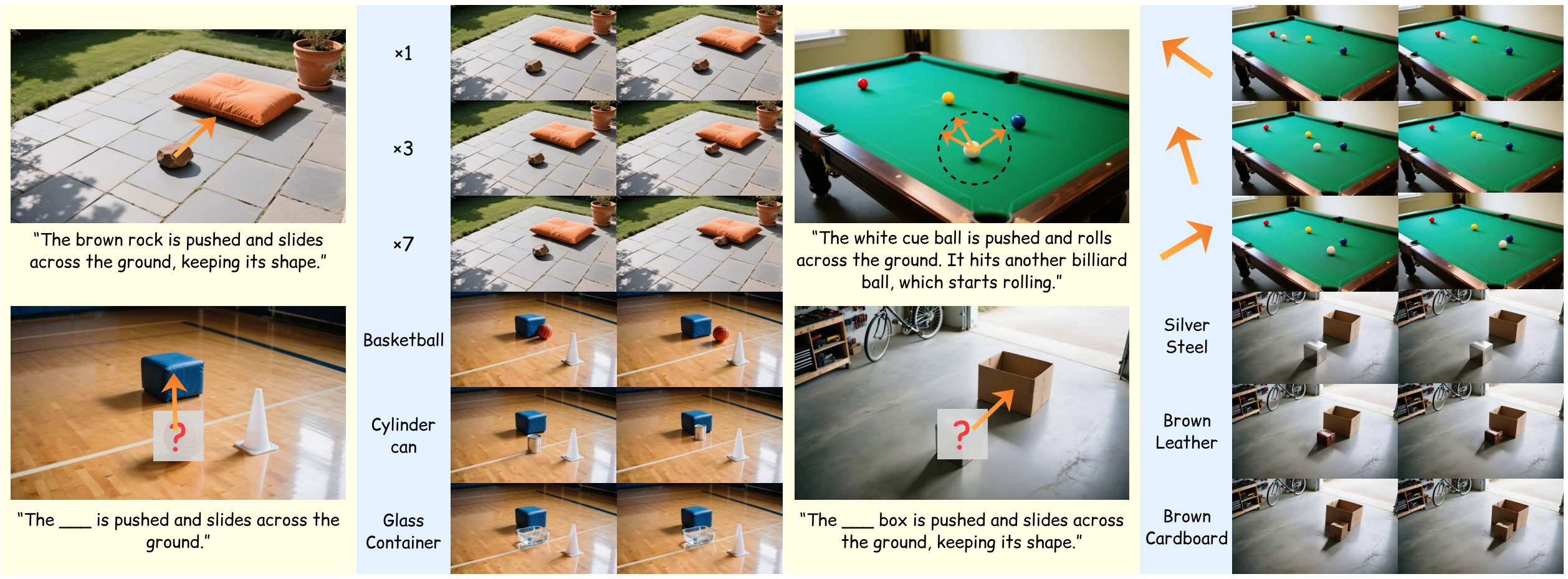}
  \caption{\textbf{Physics awareness.} Top left: external interaction intensity ($1\times$, $3\times$, $7\times$) applied to the same rock. Top right: external interaction direction applied to the same cue ball. Bottom right: material (steel, leather, cardboard) under the same push. Bottom left: shape (basketball, can, glass container) under the same push.}
  \label{fig:force_shape}
\end{figure*}

We evaluate our approach along three axes: its physical accuracy compared with methods that learn forces or collisions directly (\cref{sec:exp:accuracy}), its physics awareness (\cref{sec:exp:awareness}), and the contribution of each component (\cref{sec:ablation}).

\subsection{Experimental Setup}
\label{sec:exp:setup}

\paragraph{Implementation details.}
We train at $480\times832$ resolution with the three stages of \cref{sec:method:training}; hyperparameters are in the Appendix. Depth, segmentation, and point tracks come from Depth-Anything-3~\cite{da3}, SAM2~\cite{sam2}, and CoTracker3~\cite{cotracker3}, and the controller queries Qwen2.5-VL-3B~\cite{qwen25vl} for object properties. More details are provided in Appendix.

\paragraph{Benchmarks.}
\emph{Physics-IQ}~\cite{physicsiq} compares generated videos with real
recordings of physical events. We use its \emph{solid-mechanics} subset,
the rigid-body scenarios our controls address (\eg, collisions, rolling,
sliding, toppling). All methods
use the same subset and conditioning frame.
Since Physics-IQ has few material contrasts, we also build a
\emph{real-image test set} of 200 images in groups that share the scene
and push and differ only in the material of the pushed object (\eg,
rubber \vs steel), so a method cannot succeed by producing the same behavior for every material.

\paragraph{Baselines.}
We compare against Wan2.2-TI2V-5B~\cite{wan22}, which receives each push
as text appended to the prompt;
Force Prompting~\cite{gillman2025forceprompting}, a bidirectional model
conditioned on a first-frame force; Goal Force~\cite{goalforce}, trained on
synthetic ball and domino collisions and evaluated in its direct-force
mode; and StreamForce~\cite{wang2026streamforce}, a causal generator
conditioned on time-varying forces. For pushes issued during streaming, we
also compare qualitatively with RealWonder~\cite{liu2026realwonder}, which
simulates reconstructed 3D scenes, and FlashMotion~\cite{li2026flashmotion},
which follows trajectories.

\paragraph{Metrics.}
On Physics-IQ, we report the Physics-IQ score (P-IQ), which aggregates the
benchmark's metrics normalized so that two real recordings of the same
event score 100, under the verified and original protocols, together with
its spatiotemporal (ST), spatial (S), and weighted spatial (WS) IoU
components, averaged over 5 seeds.
Since real images lack ground-truth videos, we follow the VLM-based
protocol of VideoPhy~\cite{videophy,wang2025physctrl} and query GPT-4o for
semantic adherence (SA) and physical commonsense (PC) on a 1--5 Likert
scale. Push success (the share of pushes that set
the object in motion) measures responsiveness, and the median direction
error measures whether it moves the right way.

\subsection{Comparison with Baselines}
\label{sec:exp:accuracy}

\paragraph{Quantitative Comparison.}
On Physics-IQ, WorldInteract achieves the highest score under both protocols
(\cref{tab:phyiq}), 15 points above the strongest baseline, Goal Force,
even though Goal Force uses a much larger backbone.
The IoU components explain the gain. On spatial IoU, which checks only \emph{where} motion happens, Goal Force
is close to our model. On spatiotemporal IoU, which
also checks \emph{when} it happens, our model scores more than twice as high as any baseline. Both models move objects to similar places, but ours moves them at the right time.

On real-world images (\cref{tab:realimg}), our model raises push success from
67\% (StreamForce, the strongest baseline) to 90\%, reduces the median
direction error from 10.1$^\circ$ to 5.2$^\circ$, and obtains the highest SA and PC. The baselines instead trade these criteria off: Wan2.2 attains a high SA but moves the object in only 30\% of pushes, whereas Force Prompting achieves a high PC at the cost of the lowest SA.

\paragraph{Qualitative comparison.}
Figure~\ref{fig:qualitative_nonstream} shows that the baselines fail in three
characteristic ways. First, Force Prompting and StreamForce often barely
move the pushed object, so the described contact never happens (\eg, the
tissue box does not reach the pillow). Second, Goal Force moves objects but
violates rigid-body behavior: the rock breaks into fragments, tissues spill out of the tissue box, and the snow heap bursts into a spray. Third, Wan2.2 adds content instead of simulating the push, for instance a hand that enters the food-container scene, which explains its high SA but low PC. Our model keeps each pushed object intact and resolves the contact
the prompt describes: the rock stops against the heavy boulder, the tissue
box slides into the pillow, and the first food container pushes the second. In streaming rollouts
(\cref{fig:qualitative_stream}), RealWonder distorts pushed objects,
FlashMotion leaves them static, and StreamForce loses scene detail,
whereas WorldInteract resolves the glass--can contact.

\begin{table}[!t]
  \centering\resulttablestyle
  \caption{\textbf{Physics-IQ evaluation.}~\cite{physicsiq} Physics-IQ (P-IQ) scores use the verified and original protocols. IoU components are spatiotemporal (ST), spatial (S), and weighted spatial (WS). Best results are bold with a pale blue background.}
  \label{tab:phyiq}
  \begin{tabular*}{\linewidth}{@{\extracolsep{\fill}\hspace{\tabcolsep}}cccccc}
    \toprule
    \multirow{2}{*}{Method} & \multicolumn{2}{c}{P-IQ $\uparrow$} & \multicolumn{3}{c}{IoU $\uparrow$}\\
    \cmidrule(lr){2-3}\cmidrule(lr){4-6}
    & Verified & Original & ST & S & WS\\
    \midrule
    Wan2.2 (text-guided) & 31.5 & 24.6 & 0.338 & 0.308 & 0.264\\
    Force Prompting & 34.7 & 30.4 & 0.095 & 0.533 & 0.395\\
    StreamForce & 30.8 & 25.4 & 0.071 & 0.474 & 0.372\\
    Goal Force & 48.0 & 52.9 & 0.280 & 0.745 & 0.475\\
    \midrule
    \textbf{WorldInteract} & \bestresult{63.0} & \bestresult{68.2} & \bestresult{0.701} &\bestresult{0.779}& \bestresult{0.607}\\
    \bottomrule
  \end{tabular*}
\end{table}

\begin{table}[!t]
  \centering\resulttablestyle
  \caption{\textbf{Evaluation on real-world images.} Semantic adherence (SA) and physical commonsense (PC) are rated by a VLM judge on a 1--5 scale. Push success is the percentage of pushes that set the target in motion; direction error is the median absolute angle between its tracked motion and the push. Best results are bold with a pale blue background.}
  \label{tab:realimg}
  \begin{tabular*}{\linewidth}{@{\extracolsep{\fill}\hspace{\tabcolsep}}ccccc}
    \toprule
    \multirow{2}{*}{Method} & \multicolumn{2}{c}{Quality $\uparrow$} & \multicolumn{2}{c}{Push control}\\
    \cmidrule(lr){2-3}\cmidrule(lr){4-5}
    & SA & PC & Success (\%) $\uparrow$ & Dir.\ error ($^\circ$) $\downarrow$\\
    \midrule
    Wan2.2 (text-guided) & 4.55 & 2.93 & 30 & 36.6\\
    Force Prompting & 3.54 & 4.15 & 53 & 16.4\\
    StreamForce & 3.69 & 4.05 & 67 & 10.1\\
    Goal Force & 4.76 & 3.65 & 60 & 18.9\\
    \midrule
    \textbf{WorldInteract} & \bestresult{4.85} & \bestresult{4.30} & \bestresult{90} & \bestresult{5.2}\\
    \bottomrule
  \end{tabular*}
\end{table}

\subsection{Physics Awareness}
\label{sec:exp:awareness}
\label{sec:exp:material}

Figure~\ref{fig:force_shape} varies one physical factor at a time while holding
the others fixed. \emph{External interaction intensity}: the rock barely
moves under the $1\times$ push, travels farther under $3\times$, and reaches
the cushion under $7\times$. \emph{External interaction direction}: changing
the push direction steers the cue ball toward different billiard balls,
transferring motion upon contact. \emph{Material}: under the same push,
the steel box barely moves, the leather box travels a short distance, and
the cardboard box reaches the large box, consistent with their estimated
masses. \emph{Shape}: the basketball rolls into the pouf and rebounds,
the can reaches it more slowly, and the container slides and stops short,
showing distinct rolling and sliding behaviors. The Appendix provides
baseline comparisons.

\subsection{Ablation Study}
\label{sec:ablation}

\paragraph{Component ablations.}
Each component contributes to the final model (\cref{tab:stage_ablation}).
Causal state propagation matters most: without Stages~2 and~3, P-IQ falls
from 63.0 to 52.4 and push success from 90\% to 82\%, highlighting the importance of
tracking objects as the scene evolves. Material-Aware RL adds a
further 2.0 P-IQ points on top of supervised training. The physics
controller leaves push success unchanged but adds 4.0 P-IQ points,
indicating that explicit contact computation drives the gain in
physical accuracy.

\begin{table}[!t]
  \centering\resulttablestyle
  \caption{\textbf{Ablation study.} Variants omit the indicated training stages or the physics controller. P-IQ uses the verified protocol; PC and push-control metrics follow Table~\ref{tab:realimg}. Best results are bold with a pale blue background.}
  \label{tab:stage_ablation}
  \begin{tabular*}{\linewidth}{@{\extracolsep{\fill}\hspace{\tabcolsep}}ccccc}
    \toprule
    \multirow{2}{*}{Variant} & P-IQ $\uparrow$ & \multirow{2}{*}{PC $\uparrow$} & \multicolumn{2}{c}{Push control}\\
    \cmidrule(lr){2-2}\cmidrule(lr){4-5}
    & Verified & & Success (\%) $\uparrow$ & Dir.\ error ($^\circ$) $\downarrow$\\
    \midrule
    w/o Stages 2 \& 3 & 52.4 & 3.8 & 82 & 8.0\\
    w/o Stage 3 & 61.0 & 4.2 & 86 & 5.4\\
    w/o physics controller & 59.0 & 4.0 & \bestresult{90} & 5.3\\
    \midrule
    \textbf{WorldInteract} & \bestresult{63.0} & \bestresult{4.3} & \bestresult{90} & \bestresult{5.2}\\
    \bottomrule
  \end{tabular*}
\end{table}

\section{Conclusion}
\label{sec:conclusion}
We presented WorldInteract, a streaming video generation framework that
combines material-dependent behavior generation with contact response
computation through a shared
velocity-increment interface. Interaction-free videos teach reusable material behavior, causal training with scene memory maintains spatial state and
object identity, and Material-Aware RL improves the execution of physical
controls on generated rollouts. A physics controller composes contact
responses from object properties estimated by a VLM, allowing supervised
data to focus on individual materials without enumerating every material
pairing and contact configuration. The existing evaluations on Physics-IQ
and real images support the effectiveness of this decomposition for
physical accuracy and control adherence. The current controller uses
approximate geometry and a frictionless central collision model; extending
the same interface to richer contact models is a direction for future work.

\section*{LLM Usage Statement}
LLMs played a supporting role, helping with selected implementation and experimental coding tasks, debugging, feedback on experimental plans, and literature searches. The authors take responsibility for the final manuscript and reported results.

{\small
  \bibliographystyle{ieeenat_fullname}
  \bibliography{main,refs}

\begin{thebibliography}{37}
\providecommand{\natexlab}[1]{#1}
\providecommand{\url}[1]{\texttt{#1}}
\expandafter\ifx\csname urlstyle\endcsname\relax
  \providecommand{\doi}[1]{doi: #1}\else
  \providecommand{\doi}{doi: \begingroup \urlstyle{rm}\Url}\fi

\bibitem[Bai et~al.(2025)Bai, Chen, Liu, Wang, Ge, Song, Dang, Wang, Wang, Tang, Zhong, Zhu, Yang, Li, Wan, Wang, Ding, Fu, Xu, Ye, Zhang, Xie, Cheng, Zhang, Yang, Xu, and Lin]{qwen25vl}
Shuai Bai, Keqin Chen, Xuejing Liu, Jialin Wang, Wenbin Ge, Sibo Song, Kai Dang, Peng Wang, Shijie Wang, Jun Tang, Humen Zhong, Yuanzhi Zhu, Mingkun Yang, Zhaohai Li, Jianqiang Wan, Pengfei Wang, Wei Ding, Zheren Fu, Yiheng Xu, Jiabo Ye, Xi Zhang, Tianbao Xie, Zesen Cheng, Hang Zhang, Zhibo Yang, Haiyang Xu, and Junyang Lin.
\newblock {Qwen2.5-VL} technical report.
\newblock \emph{arXiv preprint arXiv:2502.13923}, 2025.

\bibitem[Bansal et~al.(2025)Bansal, Lin, Xie, Zong, Yarom, Bitton, Jiang, Sun, Chang, and Grover]{videophy}
Hritik Bansal, Zongyu Lin, Tianyi Xie, Zeshun Zong, Michal Yarom, Yonatan Bitton, Chenfanfu Jiang, Yizhou Sun, Kai-Wei Chang, and Aditya Grover.
\newblock Videophy: Evaluating physical commonsense for video generation.
\newblock In \emph{ICLR}, 2025.

\bibitem[Chen et~al.(2024)Chen, Mart{\'i}~Mons{\'o}, Du, Simchowitz, Tedrake, and Sitzmann]{chen2024diffusionforcing}
Boyuan Chen, Diego Mart{\'i}~Mons{\'o}, Yilun Du, Max Simchowitz, Russ Tedrake, and Vincent Sitzmann.
\newblock Diffusion forcing: Next-token prediction meets full-sequence diffusion.
\newblock In \emph{NeurIPS}, 2024.

\bibitem[Chen et~al.(2025)Chen, Jiang, Liu, Gupta, Li, Zhao, and Wang]{chen2025physgen3d}
Boyuan Chen, Hanxiao Jiang, Shaowei Liu, Saurabh Gupta, Yunzhu Li, Hao Zhao, and Shenlong Wang.
\newblock {PhysGen3D: Crafting a Miniature Interactive World from a Single Image}.
\newblock In \emph{CVPR}, 2025.

\bibitem[Chen et~al.(2026)Chen, Wonka, Wang, Wang, Feng, Tulyakov, and Liu]{chen2026physstream}
Chuhao Chen, Peter Wonka, Chaoyang Wang, Chen Wang, Qiao Feng, Sergey Tulyakov, and Lingjie Liu.
\newblock {PhysStream: Streaming Physics-Grounded Video Generation with Structured Scene Memory and Fine-Grained Motion Control}.
\newblock In \emph{SIGGRAPH Asia}, 2026.

\bibitem[Coumans and Bai(2016--2021)]{pybullet}
Erwin Coumans and Yunfei Bai.
\newblock {PyBullet}, a {Python} module for physics simulation for games, robotics and machine learning.
\newblock \url{http://pybullet.org}, 2016--2021.

\bibitem[Foo et~al.(2026)Foo, Huang, Lattas, Moschoglou, Beeler, and Theobalt]{foo2026psivg}
Lin~Geng Foo, Mark~He Huang, Alexandros Lattas, Stylianos Moschoglou, Thabo Beeler, and Christian Theobalt.
\newblock {Physical Simulator In-the-Loop Video Generation}.
\newblock In \emph{CVPR}, 2026.

\bibitem[Gillman et~al.(2025)Gillman, Herrmann, Freeman, Aggarwal, Luo, Sun, and Sun]{gillman2025forceprompting}
Nate Gillman, Charles Herrmann, Michael Freeman, Daksh Aggarwal, Evan Luo, Deqing Sun, and Chen Sun.
\newblock {Force Prompting: Video Generation Models Can Learn and Generalize Physics-based Control Signals}.
\newblock In \emph{NeurIPS}, 2025.

\bibitem[Gillman et~al.(2026)Gillman, Zhou, Tang, Luo, Chakravarthy, Aggarwal, Freeman, Herrmann, and Sun]{goalforce}
Nate Gillman, Yinghua Zhou, Zitian Tang, Evan Luo, Arjan Chakravarthy, Daksh Aggarwal, Michael Freeman, Charles Herrmann, and Chen Sun.
\newblock {Goal Force}: Teaching video models to accomplish physics-conditioned goals.
\newblock In \emph{CVPR}, 2026.

\bibitem[Henschel et~al.(2025)Henschel, Khachatryan, Poghosyan, Hayrapetyan, Tadevosyan, Wang, Navasardyan, and Shi]{henschel2024streamingt2v}
Roberto Henschel, Levon Khachatryan, Hayk Poghosyan, Daniil Hayrapetyan, Vahram Tadevosyan, Zhangyang Wang, Shant Navasardyan, and Humphrey Shi.
\newblock {StreamingT2V: Consistent, Dynamic, and Extendable Long Video Generation from Text}.
\newblock In \emph{CVPR}, 2025.

\bibitem[Huang et~al.(2025)Huang, Li, He, Zhou, and Shechtman]{huang2025selfforcing}
Xun Huang, Zhengqi Li, Guande He, Mingyuan Zhou, and Eli Shechtman.
\newblock Self forcing: Bridging the train-test gap in autoregressive video diffusion.
\newblock In \emph{NeurIPS}, 2025.

\bibitem[Karaev et~al.(2025)Karaev, Makarov, Wang, Neverova, Vedaldi, and Rupprecht]{cotracker3}
Nikita Karaev, Iurii Makarov, Jianyuan Wang, Natalia Neverova, Andrea Vedaldi, and Christian Rupprecht.
\newblock {CoTracker3}: Simpler and better point tracking by pseudo-labelling real videos.
\newblock In \emph{ICCV}, 2025.

\bibitem[Kim et~al.(2024)Kim, Kang, Choi, and Han]{kim2024fifodiffusion}
Jihwan Kim, Junoh Kang, Jinyoung Choi, and Bohyung Han.
\newblock {FIFO-Diffusion: Generating Infinite Videos from Text without Training}.
\newblock In \emph{NeurIPS}, 2024.

\bibitem[Li et~al.(2026{\natexlab{a}})Li, Liu, Lin, and Chandraker]{li2026rollingsink}
Haodong Li, Shaoteng Liu, Zhe Lin, and Manmohan Chandraker.
\newblock {Rolling Sink: Bridging Limited-Horizon Training and Open-Ended Testing in Autoregressive Video Diffusion}.
\newblock \emph{arXiv preprint arXiv:2602.07775}, 2026{\natexlab{a}}.

\bibitem[Li et~al.(2026{\natexlab{b}})Li, Xing, Wang, Cao, Dai, Dong, and Wu]{li2026flashmotion}
Quanhao Li, Zhen Xing, Rui Wang, Haidong Cao, Qi Dai, Daoguo Dong, and Zuxuan Wu.
\newblock {FlashMotion: Few-Step Controllable Video Generation with Trajectory Guidance}.
\newblock In \emph{CVPR}, 2026{\natexlab{b}}.

\bibitem[Lin et~al.(2026)Lin, Chen, Liew, Chen, Li, Shi, Feng, and Kang]{da3}
Haotong Lin, Sili Chen, Jun~Hao Liew, Donny~Y. Chen, Zhenyu Li, Guang Shi, Jiashi Feng, and Bingyi Kang.
\newblock {Depth Anything 3}: Recovering the visual space from any views.
\newblock In \emph{ICLR}, 2026.

\bibitem[Lipman et~al.(2023)Lipman, Chen, Ben-Hamu, Nickel, and Le]{flowmatching}
Yaron Lipman, Ricky T.~Q. Chen, Heli Ben-Hamu, Maximilian Nickel, and Matt Le.
\newblock Flow matching for generative modeling.
\newblock In \emph{ICLR}, 2023.

\bibitem[Liu et~al.(2026{\natexlab{a}})Liu, Hu, Xu, Shan, and Lu]{liu2025rollingforcing}
Kunhao Liu, Wenbo Hu, Jiale Xu, Ying Shan, and Shijian Lu.
\newblock {Rolling Forcing: Autoregressive Long Video Diffusion in Real Time}.
\newblock In \emph{ICLR}, 2026{\natexlab{a}}.

\bibitem[Liu et~al.(2024)Liu, Ren, Gupta, and Wang]{liu2024physgen}
Shaowei Liu, Zhongzheng Ren, Saurabh Gupta, and Shenlong Wang.
\newblock {PhysGen: Rigid-Body Physics-Grounded Image-to-Video Generation}.
\newblock In \emph{ECCV}, 2024.

\bibitem[Liu et~al.(2026{\natexlab{b}})Liu, Chen, Li, Wang, Yu, and Wu]{liu2026realwonder}
Wei Liu, Ziyu Chen, Zizhang Li, Yue Wang, Hong-Xing Yu, and Jiajun Wu.
\newblock {RealWonder: Real-Time Physical Action-Conditioned Video Generation}.
\newblock \emph{arXiv preprint arXiv:2603.05449}, 2026{\natexlab{b}}.

\bibitem[Motamed et~al.(2026)Motamed, Culp, Swersky, Jaini, and Geirhos]{physicsiq}
Saman Motamed, Laura Culp, Kevin Swersky, Priyank Jaini, and Robert Geirhos.
\newblock Do generative video models understand physical principles?
\newblock In \emph{WACV}, 2026.

\bibitem[Rafailov et~al.(2023)Rafailov, Sharma, Mitchell, Manning, Ermon, and Finn]{dpo}
Rafael Rafailov, Archit Sharma, Eric Mitchell, Christopher~D. Manning, Stefano Ermon, and Chelsea Finn.
\newblock Direct preference optimization: Your language model is secretly a reward model.
\newblock In \emph{NeurIPS}, 2023.

\bibitem[Ravi et~al.(2025)Ravi, Gabeur, Hu, Hu, Ryali, Ma, Khedr, R{\"a}dle, Rolland, Gustafson, Mintun, Pan, Alwala, Carion, Wu, Girshick, Doll{\'a}r, and Feichtenhofer]{sam2}
Nikhila Ravi, Valentin Gabeur, Yuan-Ting Hu, Ronghang Hu, Chaitanya Ryali, Tengyu Ma, Haitham Khedr, Roman R{\"a}dle, Chloe Rolland, Laura Gustafson, Eric Mintun, Junting Pan, Kalyan~Vasudev Alwala, Nicolas Carion, Chao-Yuan Wu, Ross Girshick, Piotr Doll{\'a}r, and Christoph Feichtenhofer.
\newblock {SAM} 2: Segment anything in images and videos.
\newblock In \emph{ICLR}, 2025.

\bibitem[Shin et~al.(2026)Shin, Li, Zhang, Zhu, Park, Shechtman, and Huang]{shin2025motionstream}
Joonghyuk Shin, Zhengqi Li, Richard Zhang, Jun-Yan Zhu, Jaesik Park, Eli Shechtman, and Xun Huang.
\newblock {MotionStream: Real-Time Video Generation with Interactive Motion Controls}.
\newblock In \emph{ICLR}, 2026.

\bibitem[Song et~al.(2025)Song, Chen, Simchowitz, Du, Tedrake, and Sitzmann]{song2025historyguided}
Kiwhan Song, Boyuan Chen, Max Simchowitz, Yilun Du, Russ Tedrake, and Vincent Sitzmann.
\newblock {History-Guided Video Diffusion}.
\newblock In \emph{ICML}, 2025.

\bibitem[Tan et~al.(2026)Tan, Jiang, Li, Zong, Xie, Yang, and Jiang]{tan2024physmotion}
Xiyang Tan, Ying Jiang, Xuan Li, Zeshun Zong, Tianyi Xie, Yin Yang, and Chenfanfu Jiang.
\newblock {PhysMotion: Physics-Grounded Dynamics From a Single Image}.
\newblock In \emph{3DV}, 2026.

\bibitem[{Team Wan} et~al.(2025){Team Wan}, Wang, Ai, Wen, Mao, Xie, Chen, Yu, Zhao, Yang, Zeng, Wang, Zhang, Zhou, Wang, Chen, Zhu, Zhao, Yan, Huang, Feng, Zhang, Li, Wu, Chu, Feng, Zhang, Sun, Fang, Wang, Gui, Weng, Shen, Lin, Wang, Wang, Zhou, Wang, Shen, Yu, Shi, Huang, Xu, Kou, Lv, Li, Liu, Wang, Zhang, Huang, Li, Wu, Liu, Pan, Zheng, Hong, Shi, Feng, Jiang, Han, Wu, and Liu]{wan22}
{Team Wan}, Ang Wang, Baole Ai, Bin Wen, Chaojie Mao, Chen-Wei Xie, Di Chen, Feiwu Yu, Haiming Zhao, Jianxiao Yang, Jianyuan Zeng, Jiayu Wang, Jingfeng Zhang, Jingren Zhou, Jinkai Wang, Jixuan Chen, Kai Zhu, Kang Zhao, Keyu Yan, Lianghua Huang, Mengyang Feng, Ningyi Zhang, Pandeng Li, Pingyu Wu, Ruihang Chu, Ruili Feng, Shiwei Zhang, Siyang Sun, Tao Fang, Tianxing Wang, Tianyi Gui, Tingyu Weng, Tong Shen, Wei Lin, Wei Wang, Wei Wang, Wenmeng Zhou, Wente Wang, Wenting Shen, Wenyuan Yu, Xianzhong Shi, Xiaoming Huang, Xin Xu, Yan Kou, Yangyu Lv, Yifei Li, Yijing Liu, Yiming Wang, Yingya Zhang, Yitong Huang, Yong Li, You Wu, Yu Liu, Yulin Pan, Yun Zheng, Yuntao Hong, Yupeng Shi, Yutong Feng, Zeyinzi Jiang, Zhen Han, Zhi-Fan Wu, and Ziyu Liu.
\newblock Wan: Open and advanced large-scale video generative models.
\newblock \emph{arXiv preprint arXiv:2503.20314}, 2025.

\bibitem[Wallace et~al.(2024)Wallace, Dang, Rafailov, Zhou, Lou, Purushwalkam, Ermon, Xiong, Joty, and Naik]{diffusiondpo}
Bram Wallace, Meihua Dang, Rafael Rafailov, Linqi Zhou, Aaron Lou, Senthil Purushwalkam, Stefano Ermon, Caiming Xiong, Shafiq Joty, and Nikhil Naik.
\newblock Diffusion model alignment using direct preference optimization.
\newblock In \emph{CVPR}, 2024.

\bibitem[Wang et~al.(2025)Wang, Chen, Huang, Dou, Liu, Gu, and Liu]{wang2025physctrl}
Chen Wang, Chuhao Chen, Yiming Huang, Zhiyang Dou, Yuan Liu, Jiatao Gu, and Lingjie Liu.
\newblock {PhysCtrl: Generative Physics for Controllable and Physics-Grounded Video Generation}.
\newblock In \emph{NeurIPS}, 2025.

\bibitem[Wang et~al.(2026)Wang, Xie, Feng, Lv, Wang, and Jiang]{wang2026streamforce}
Hanhui Wang, Yiming Xie, Haiwen Feng, Zhaoyang Lv, Shenlong Wang, and Huaizu Jiang.
\newblock {Streaming Video Generation with Streaming Force Control}.
\newblock In \emph{{SIGGRAPH} Asia}, 2026.

\bibitem[Wu et~al.(2024)Wu, Li, Gu, Zhao, He, Zhang, Shou, Li, Gao, and Zhang]{wu2024draganything}
Weijia Wu, Zhuang Li, Yuchao Gu, Rui Zhao, Yefei He, David~Junhao Zhang, Mike~Zheng Shou, Yan Li, Tingting Gao, and Di Zhang.
\newblock {DragAnything: Motion Control for Anything using Entity Representation}.
\newblock In \emph{ECCV}, 2024.

\bibitem[Yang et~al.(2026)Yang, Huang, Chu, Xiao, Zhao, Wang, Li, Xie, Chen, Lu, Han, and Chen]{yang2025longlive}
Shuai Yang, Wei Huang, Ruihang Chu, Yicheng Xiao, Yuyang Zhao, Xianbang Wang, Muyang Li, Enze Xie, Yingcong Chen, Yao Lu, Song Han, and Yukang Chen.
\newblock {LongLive: Real-time Interactive Long Video Generation}.
\newblock In \emph{ICLR}, 2026.

\bibitem[Yin et~al.(2024)Yin, Gharbi, Zhang, Shechtman, Durand, Freeman, and Park]{dmd}
Tianwei Yin, Micha{\"e}l Gharbi, Richard Zhang, Eli Shechtman, Fr{\'e}do Durand, William~T. Freeman, and Taesung Park.
\newblock One-step diffusion with distribution matching distillation.
\newblock In \emph{CVPR}, 2024.

\bibitem[Yin et~al.(2025)Yin, Zhang, Zhang, Freeman, Durand, Shechtman, and Huang]{yin2024causvid}
Tianwei Yin, Qiang Zhang, Richard Zhang, William~T Freeman, Fredo Durand, Eli Shechtman, and Xun Huang.
\newblock {From Slow Bidirectional to Fast Autoregressive Video Diffusion Models}.
\newblock In \emph{CVPR}, 2025.

\bibitem[Zhang et~al.(2025)Zhang, Liao, Li, Dai, Qiu, Zhu, Qin, and Wang]{zhang2024tora}
Zhenghao Zhang, Junchao Liao, Menghao Li, Zuozhuo Dai, Bingxue Qiu, Siyu Zhu, Long Qin, and Weizhi Wang.
\newblock {Tora: Trajectory-oriented Diffusion Transformer for Video Generation}.
\newblock In \emph{CVPR}, 2025.

\bibitem[Zhou et~al.(2026)Zhou, Zhou, Zhao, Xu, Zhu, Hong, and Zhang]{zhou2025dragstream}
Junbao Zhou, Yuan Zhou, Kesen Zhao, Qingshan Xu, Beier Zhu, Richang Hong, and Hanwang Zhang.
\newblock {Streaming Drag-Oriented Interactive Video Manipulation: Drag Anything, Anytime!}
\newblock In \emph{ICLR}, 2026.

\bibitem[Zhu et~al.(2026)Zhu, Zhao, He, Su, Li, and Zhu]{causalforcing}
Hongzhou Zhu, Min Zhao, Guande He, Hang Su, Chongxuan Li, and Jun Zhu.
\newblock Causal forcing: Autoregressive diffusion distillation done right for high-quality real-time interactive video generation.
\newblock In \emph{ICML}, 2026.

\end{thebibliography}
}

\clearpage
\appendix
\setcounter{equation}{0}
\setcounter{figure}{0}
\setcounter{table}{0}
\setcounter{algorithm}{0}
\renewcommand{\theequation}{S\arabic{equation}}
\renewcommand{\thefigure}{S\arabic{figure}}
\renewcommand{\thetable}{S\arabic{table}}
\renewcommand{\thealgorithm}{S\arabic{algorithm}}
\renewcommand{\theHequation}{appendix.\arabic{equation}}
\renewcommand{\theHfigure}{appendix.\arabic{figure}}
\renewcommand{\theHtable}{appendix.\arabic{table}}
\def\theHalgorithm{appendix.\arabic{algorithm}}
\twocolumn[{%
  \centering\Large\textbf{Appendix}\par\vspace{1em}
}]

This appendix complements the main paper. \Cref{sec:supp:data} describes
the interaction-free dataset. \Cref{sec:supp:implementation} details the
implementation and training settings, while
\cref{sec:supp:causal,sec:supp:objectives} explain causal generation and
the supervised objectives. \Cref{sec:supp:rl} presents Material-Aware RL,
\cref{sec:supp:inference} describes the physics controller, and
\cref{sec:supp:qualitative} provides additional qualitative results.

\section{Interaction-free Dataset}
\label{sec:supp:data}
Each clip shows one simulated object pushed horizontally on a textured
floor under an HDRI environment map, rendered with Blender
Cycles~\cite{blender} at $480\times832$ and 16\,fps. Training uses 49
frames per clip. Other objects in the scene either stay still or move in
their own separate simulation, so no object--object contact occurs.
Contact with the floor, friction, and rebounds remain. The 30k clips
span 22 material classes across rigid and elastic solids, sand, and fluids
(\cref{tab:supp:dataset,tab:supp:materials}).

\begin{table*}[t]
  \centering\footnotesize
  \renewcommand{\arraystretch}{1.25}
  \setlength{\tabcolsep}{4pt}
  \caption{\textbf{Interaction-free dataset.} All clips are rendered with Blender Cycles at $480\times832$ and 16\,fps. Pushes are horizontal. Shares are fractions of the 30k clips.}
  \label{tab:supp:dataset}
  \begin{tabularx}{\textwidth}{@{}>{\raggedright\arraybackslash}p{1.1cm}>{\raggedright\arraybackslash}p{3.6cm}>{\raggedright\arraybackslash}X>{\raggedright\arraybackslash}p{4.4cm}r@{}}
    \toprule
    Category & Simulation & Objects and materials & Push & Share\\
    \midrule
    Rigid
    & PyBullet~\cite{pybullet} rigid body
    & 2,200 procedural shapes with base masses of 0.5--3\,kg and scanned or procedural textures. 15 materials (\cref{tab:supp:materials}), each with its own density, friction, and restitution.
    & 0.1\,s force of 0.3--35\,N at a random side point, within $\pm60^\circ$ of the surface normal; 30\% of clips receive 2--3 pushes & 67.3\%\\
    \addlinespace
    Elastic
    & MLS-MPM~\cite{hu2018mlsmpm} in NVIDIA Warp~\cite{warp}, Neo-Hookean
    & The same shapes as soft bodies (Young's modulus 4.5\,kPa, Poisson ratio 0.25--0.40). 5 materials (\cref{tab:supp:materials}).
    & As for rigid objects, with the force spread over a Gaussian patch & 28.4\%\\
    \addlinespace
    Sand
    & MPM, Drucker--Prager~\cite{klar2016sand}
    & A sand heap on the floor
    & Part of the heap is pushed so that it slides and spreads & 2.15\%\\
    \addlinespace
    Fluid
    & MPM with the Tait equation of state~\cite{becker2007wcsph}
    & Water in a cup or jar, mostly glass or plastic
    & The vessel is pushed and slides while the water sloshes & 2.15\%\\
    \bottomrule
  \end{tabularx}
\end{table*}

\begin{table}[t]
  \centering\footnotesize
  \renewcommand{\arraystretch}{1.08}
  \caption{\textbf{Material classes.} Share of the 30k clips for each of the 22 material classes.}
  \label{tab:supp:materials}
  \begin{tabular*}{\linewidth}{@{\extracolsep{\fill}}llr@{\hspace{12pt}}llr@{}}
    \toprule
    Material & Type & Share & Material & Type & Share\\
    \midrule
    Fabric & Elastic & 15.60\% & Terrazzo & Rigid & 3.80\%\\
    Vinyl & Elastic & 6.60\% & Cardboard & Rigid & 3.66\%\\
    Leather & Elastic & 4.30\% & Concrete & Rigid & 3.66\%\\
    Carpet & Elastic & 1.20\% & Glass & Rigid & 3.46\%\\
    Rubber & Elastic & 0.70\% & Steel & Rigid & 3.26\%\\
    Sand & Granular & 2.15\% & Foam & Rigid & 3.26\%\\
    Water & Fluid & 2.15\% & Marble & Rigid & 2.90\%\\
    Plastic & Rigid & 14.20\% & Onyx & Rigid & 2.40\%\\
    Wood & Rigid & 8.80\% & Porcelain & Rigid & 1.80\%\\
    Ceramic & Rigid & 8.60\% & Travertine & Rigid & 1.30\%\\
    Metal & Rigid & 5.40\% & Pine & Rigid & 0.80\%\\
    \bottomrule
  \end{tabular*}
\end{table}

\paragraph{Pushes.}
A rigid or elastic push applies a horizontal force for 0.1\,s; its impulse
$\mathbf J$ is the force times this duration, and the simulator's true mass
gives the velocity increment $\Delta\mathbf v=\mathbf J/m_o$. In clips with
several pushes, the later pushes act during motion, including re-pushes of
the already moving object.

\paragraph{Paired groups.}
The rigid and elastic categories also contain clips rendered in
\emph{paired groups} that share the scene. \emph{Force ladders} repeat a
scene at $0.4$, $0.6$, $0.8$, and $1.0\times$ the push force, and
\emph{material pairs} change the pushed object's material, making it
$2$--$2.5\times$ heavier or $0.4$--$0.5\times$ lighter. These groups supply
the simulated pairs for Material-Aware RL (\cref{sec:supp:rl}). Hard
negatives, in which the pushed object passes close to other objects
without touching them, are also included.

\section{Implementation Details}
\label{sec:supp:implementation}

\paragraph{Velocity-increment encoding.}
A push $(t,o,\mathbf J)$ applies a camera-frame impulse $\mathbf J$ to
object $o$ at frame $t$. During training, the corresponding velocity
increment uses the simulator's true mass; the generator does not receive
mass as an input. Velocities are expressed in metric units in the camera
frame, using simulator units during training and metric depth from
Depth-Anything-3~\cite{da3} at inference, so $V_{\max}$ has the same
physical meaning in both. Each component is clipped to
$[-V_{\max},V_{\max}]$. The push control $c^{\mathrm u}=\{c^{\mathrm
u}_t\}_t$ is a sequence of three-channel maps. At the event frame $t$,
$c^{\mathrm u}_t=(\Delta\mathbf v/V_{\max}+\mathbf 1)/2\in[0,1]^3$ on
object $o$'s first-frame segmentation mask, where $\mathbf 1$ is the
all-ones vector; all other pixels and frames equal $0.5$, encoding a zero
increment. The first-frame mask identifies the target without revealing
its future position~\cite{chen2026physstream}.

\paragraph{Training configuration.}
\Cref{tab:supp:hparams} lists the trained and frozen modules, learning
rates, and optimizer settings of each stage. Stage~1 fine-tunes
Wan2.2-TI2V-5B~\cite{wan22} with full attention on 49-frame clips at
$480\times832$. Stage~2 starts from the Stage-1 checkpoint, and Stage~3
uses the final Stage-2 model as its frozen reference. The four-step
student is distilled from the Stage-3 model with DMD, using a critic
trained alongside it.

\begin{table*}[t]
  \centering\footnotesize
  \renewcommand{\arraystretch}{1.25}
  \setlength{\tabcolsep}{4pt}
  \caption{\textbf{Optimization settings.} All stages use AdamW with betas $(0.9, 0.999)$ unless noted.}
  \label{tab:supp:hparams}
  \begin{tabularx}{\textwidth}{@{}>{\raggedright\arraybackslash}p{2.3cm}>{\raggedright\arraybackslash}X>{\raggedright\arraybackslash}p{3.0cm}>{\raggedright\arraybackslash}p{3.4cm}>{\raggedright\arraybackslash}p{2.9cm}@{}}
    \toprule
    Stage & Trained modules & Frozen & Learning rate & Schedule\\
    \midrule
    1.~Material behavior learning
    & DiT blocks except text cross-attention K/V, time embedding and projection, video patch embedding $P_{\mathrm{pre}}$, output head, push-control branch $P_{\mathrm u}$ (4.4B)
    & Text cross-attention K/V, VAE, text encoder
    & Blocks and time projection $5{\times}10^{-5}$; $P_{\mathrm u}$ $10^{-4}$; $P_{\mathrm{pre}}$ and head $2{\times}10^{-6}$; time embedding $10^{-6}$
    & Linear warm-up, then constant; weight decay 0.01; clip 1.0\\
    \addlinespace
    2.~Causal state propagation
    & All DiT blocks incl.\ cross-attention K/V, time projection, time and text embeddings, $P_{\mathrm{pos}}$, $P_{\mathrm{trk}}$, $P_{\mathrm p}$ (5.0B)
    & $P_{\mathrm{pre}}$, $P_{\mathrm u}$, output head, VAE, text encoder
    & Blocks and embeddings $10^{-5}$; $P_{\mathrm{pos}},P_{\mathrm{trk}},P_{\mathrm p}$ $10^{-4}$
    & Constant; weight decay 0.01; clip 1.0\\
    \addlinespace
    3.~Material-Aware RL
    & Rank-64 LoRA on attention and FFN linear layers (161M); $P_{\mathrm u}$, $P_{\mathrm{pos}}$, $P_{\mathrm{trk}}$, $P_{\mathrm p}$ (2.4M)
    & All other weights
    & $10^{-5}$
    & 50-step warm-up, then constant; no weight decay; clip 1.0\\
    \addlinespace
    DMD distillation
    & Student and critic (5.0B each)
    & Stage-3 teacher
    & Student $2{\times}10^{-6}$; critic $4{\times}10^{-7}$
    & weight decay 0.01; clip 10.0; five critic updates per student update\\
    \bottomrule
  \end{tabularx}
\end{table*}

\section{Causal Generation}
\label{sec:supp:causal}

\paragraph{Causal generator.}
The causal VAE $\mathcal E$ maps a video to latents $z_{0:L}$, where $z_0$
encodes the first frame $x_0$ and each subsequent $z_\ell$ a chunk of four
frames. With block-causal attention and a key--value cache, the DiT
factorizes the video distribution as
\begin{equation}
p_\theta(z_{1:L}\mid z_0,y)=\textstyle\prod_{\ell=1}^{L} p_\theta\big(z_\ell \mid z_{<\ell},\,y,\,c_{<\ell}\big),
\label{eq:supp:factor}
\end{equation}
where $y$ is the text conditioning and $c_{<\ell}$ denotes condition maps
computed from earlier chunks only.

\paragraph{Condition maps.}
The scene memory consists of a positional map $c^{\mathrm{pos}}$,
back-projected from Depth-Anything-3~\cite{da3} depth and normalized by the
first-frame scene bounds so that positions remain comparable across chunks,
and a tracking map $c^{\mathrm{trk}}$ in which SAM2~\cite{sam2} assigns each
object a fixed color. Both are re-estimated from every decoded chunk. All
condition maps are rasterized per frame, VAE-encoded per chunk, shifted
forward by one chunk, and added to the noisy video tokens through
zero-initialized patch embeddings:
\begin{equation}
h_\ell^\tau=P_{\mathrm{pre}}\big(z^{\tau}_\ell\big)+\textstyle\sum_{m\in\mathcal M}P_m\big(\mathcal E(c^{m}_{\ell-1})\big),
\label{eq:supp:fusion}
\end{equation}
where $\mathcal M=\{\mathrm u,\mathrm{pos},\mathrm{trk},\mathrm p\}$. The
shift ensures that the conditions for chunk $\ell$ come only from chunk
$\ell-1$, so no information from chunk $\ell$ leaks into its own
conditioning. Because the VAE compresses four frames into one latent, the
timing of an event within a chunk reaches the generator only to the extent
that the encoding preserves it.

\paragraph{Physics map.}
The physics map $c^{\mathrm p}$ is encoded like $c^{\mathrm u}$ but can
carry increments for several objects and is painted on each object's
\emph{recent} mask, so the generator can identify the target after it has
moved. Because of the one-chunk shift, an increment that should act at
frame $t$ of chunk $\ell+1$ is painted into the map of chunk $\ell$, at the
same position within the chunk, on the object's mask at frame $t-4$. This
mask comes from an already generated chunk and matches the accompanying
tracking map $c^{\mathrm{trk}}_\ell$, so it does not reveal where the
object lies in the chunk being generated. The same rule is used in
training and at inference. At inference, $c^{\mathrm p}$ carries the
increments computed by the physics controller for predicted contacts.
Material therefore enters at two points: in the generator's learned
behavior (friction, rolling resistance, rebounds from the ground) and in
the controller, where mass and restitution determine the exchange at
contact (\cref{eq:supp:impulse}).

\paragraph{Physics-map supervision.}
Since the interaction-free videos contain no object--object contact, we
supervise $P_{\mathrm p}$ with ordinary pushes. For a random subset of
mid-video pushes, including all re-pushes, the increment is written to
$c^{\mathrm p}$ instead of $c^{\mathrm u}$, painted by the rule above on
the object's ground-truth mask one chunk before the push. This is the same
operation the controller performs at a contact, and because these pushes
vary in timing, magnitude, and direction, the branch learns to execute
increments of contacts it has never seen.

\paragraph{Condition perturbations.}
To narrow the gap between simulator-rendered conditions and those
estimated at inference, Stage~2 jitters the timing, magnitude, and
direction of increments and perturbs depth and tracking masks. We never
drop $c^{\mathrm p}$ while the video shows the corresponding velocity
change, since that would teach the generator to change velocities without
a cause.

\section{Supervised Objectives}
\label{sec:supp:objectives}

\paragraph{Flow matching.}
Stages~1 and~2 use rectified-flow matching~\cite{flowmatching}, following
PhysStream~\cite{chen2026physstream} with the additional physics branch in
Stage~2. Let $z_0$ encode the initial image $x_0$, $z$ denote the remaining
clean video latents, and $y$ the backbone's text conditioning. With
$\tau\sim\mathcal U(0,1)$, $\epsilon\sim\mathcal N(0,I)$, and
$z^\tau=(1-\tau)z+\tau\epsilon$, Stage~1 fuses the noisy video and push
control into input tokens
$h^\tau=P_{\mathrm{pre}}(z^\tau)+P_{\mathrm u}\big(\mathcal E(c^{\mathrm u})\big)$
and minimizes
\begin{equation}
\mathcal F_\theta(z)=\mathbb E_{\tau,\epsilon}\big\|\hat v_\theta^\tau(h^\tau,\tau,z_0,y)-(\epsilon-z)\big\|_2^2,
\label{eq:supp:fm}
\end{equation}
where $\hat v_\theta^\tau$ predicts the latent-space vector field, not an
object's physical velocity. The clean latent of $x_0$ occupies the first
slot with its condition slot set to zero, and the loss is evaluated on the
remaining latents.

\paragraph{Teacher forcing.}
Stage~2 switches to block-causal attention with one latent per
block~\cite{causalforcing} and trains with teacher forcing: each latent is
denoised given the clean ground-truth latents that precede it, with the
conditions of the preceding chunk fused into $h^\tau_\ell$ by
\cref{eq:supp:fusion},
\begin{equation}
\hat v_\ell=\hat v_\theta^\tau\big(h^{\tau}_\ell,\tau,\,z^{\mathrm{gt}}_{<\ell},\,y\big),
\label{eq:tf}
\end{equation}
and \cref{eq:supp:fm} averaged over the clip.

\section{Material-Aware RL}
\label{sec:supp:rl}

\paragraph{Motivation.}
Flow matching scores each clip in isolation and weighs all pixels equally,
so small physical errors, such as a drifting neighbor or a velocity change
one frame early, cost little, and relations such as a harder push
traveling farther are never compared. Stage~3 therefore samples rollouts
from the current model, including rollouts with controller-generated
contacts, scores them with verifiable checkers, and increases the relative
likelihood of rollouts that satisfy them. During Stage~3, the controller
uses simulator properties for simulated inputs and VLM estimates
precomputed offline for real images, so no VLM runs inside the training
loop.

\paragraph{Training inputs.}
Inputs are (i)~simulated pairs $(x_0^{+},x_0^{-})$ from the force ladders
and material pairs (\cref{sec:supp:data}), differing only in push magnitude
or in the pushed object's material, where $x_0^{+}$ travels farther in
simulation; (ii)~real-image pairs, where $x_0^{+}$ receives the stronger
push; and (iii)~\emph{contact scenes}, simulated first frames with objects
on a collision course, whose contacts occur only in the model's own
rollouts.

\paragraph{Verifiable rewards.}
Rewards are computed from SAM2 masks, CoTracker3~\cite{cotracker3} point
tracks, and depth maps. For object $k\in\mathcal O$, let $\mathbf u_k^-$
and $\mathbf u_k$ be its image velocity in the chunks before and after an
increment $\Delta\mathbf v_k$ (from $c^{\mathrm u}$ or $c^{\mathrm p}$),
$\bar{\mathbf u}_k=\mathbf u^-_k+\Pi\Delta\mathbf v_k$ its expected
velocity, with $\Pi$ the depth-based projection to the image, and $d_k$ and
$d^\star_k$ its generated and target displacements, normalized by its
first-frame bounding-box diagonal. On simulated interaction-free inputs,
$d^\star_k$ comes from the simulator. On real images and contact scenes, it
is defined only for objects that receive no increment, where $d^\star_k=0$
rewards neighbors that stay still. Objects involved in a contact are scored
only by $r_{\Delta v}$ and $r_{\mathrm{sync}}$ against the controller's
increments. Scoring them against the simulator, whose contact model
includes friction and spin, would penalize the generator for executing the
controller faithfully; Stage~3 thus teaches the generator to
\emph{execute} contact increments, while the controller remains solely
responsible for \emph{deciding} them. For a pair, let $x^+_a$ and $x^-_b$,
$a,b\in\{1,\dots,K\}$, be the rollouts of the two inputs and $d(\cdot)$ the
target's displacement. Let $t_{\mathrm m}$ be the frame at which an
increment is scheduled and $t_{\mathrm v}$ the frame at which the object's
tracked velocity changes most, found by change-point detection on its
speed. The checkers are
\begin{equation}
\begin{aligned}
r_{\mathrm{disp}}&=\tfrac{1}{|\mathcal O|}\textstyle\sum_{k\in\mathcal O}\exp\!\big(-|d_k-d^\star_k|/\rho\big),\\
r_{\Delta v}&=\tfrac{1}{|\mathcal O_\Delta|}\textstyle\sum_{k\in\mathcal O_\Delta}\exp\!\Big(-\tfrac{\|\mathbf u_k-\bar{\mathbf u}_k\|}{\max(\eta\|\bar{\mathbf u}_k\|,\,v_{\min})}\Big),\\
r_{\mathrm{rel}}(x^{+}_a)&=\tfrac1K\textstyle\sum_{b}\sigma\big((d(x^{+}_a)-d(x^{-}_b))/\kappa\big),\\
r_{\mathrm{sync}}&=\tfrac{1}{|\mathcal I|}\textstyle\sum_{(k,t_{\mathrm m})\in\mathcal I}\exp\!\big(-|t_{\mathrm v}-t_{\mathrm m}|/\omega\big),
\end{aligned}
\label{eq:checkers}
\end{equation}
where $\mathcal O_\Delta$ is the set of objects that receive an increment,
$\mathcal I$ the set of scheduled increments, $\sigma$ the logistic
function, and $\rho,\eta,v_{\min},\kappa,\omega$ temperatures;
$r_{\mathrm{rel}}(x^-_b)$ is defined
symmetrically with the sign reversed. A rollout's reward is the weighted
mean $r=\sum_c w_c r_c/\sum_c w_c$ over the checkers available for its
input ($r_{\mathrm{rel}}$ only for pairs, $r_{\mathrm{disp}}$ only for
objects with a defined $d^\star_k$).

\paragraph{Preference optimization.}
For each input, we draw $K=4$ rollouts with the 50-step sampler and different seeds. Among
rollouts with the same predicted contacts, so that the comparison is not
confounded by the controller, we take the highest- and lowest-reward ones
as winner $x^w$ and loser $x^l$, with latents $z^w$ and $z^l$, and discard
the input if their reward gap is below $0.1$. We apply
Diffusion-DPO~\cite{diffusiondpo,dpo}, adapted to flow matching, with a
frozen copy of the final Stage-2 model, $\hat v_{\mathrm{ref}}^\tau$, as
reference. Both rollouts are re-noised and denoised under teacher forcing
(\cref{eq:tf}) with their own preceding latents and condition maps. With
shared $(\tau,\epsilon)$ and
$\delta_\theta(z)=\|\hat v_\theta^\tau(h^\tau,\tau)-(\epsilon-z)\|^2-\|\hat v_{\mathrm{ref}}^\tau(h^\tau,\tau)-(\epsilon-z)\|^2$,
with conditioning omitted for brevity,
\begin{equation}
\mathcal L_{\mathrm{RL}}=-\mathbb E\,\log\sigma\big(-\beta\,[\delta_\theta(z^w)-\delta_\theta(z^l)]\big)+\lambda\,\mathcal F_\theta(z^{\mathrm{sim}}),
\label{eq:dpo}
\end{equation}
where the second term replays flow matching on simulated clips
$z^{\mathrm{sim}}$ to limit visual drift; we use $\beta=500$. We train rank-64 LoRA adapters on
the DiT's linear layers and fully fine-tune the four condition branches.

\paragraph{Acceleration.}
We distill the Stage-3 model into a four-step causal student with
DMD~\cite{dmd,yin2024causvid}. Initializing from the block-causal teacher
avoids the ODE-regression stage required when distilling a bidirectional
teacher~\cite{yin2024causvid,causalforcing}.

\section{Inference}
\label{sec:supp:inference}
Before each chunk, the physics controller updates scene memory from
committed frames, predicts upcoming contacts, and schedules the resulting
increments; the generator then executes them and commits the chunk
(\cref{alg:supp:inference}).

\begin{algorithm}[t]
  \caption{Streaming inference with the physics controller.}
  \label{alg:supp:inference}
  \small
  \begin{algorithmic}[1]
    \Require initial image $x_0$, text conditioning $y$, user pushes
    \For{each chunk $\ell=0,1,2,\dots$}
      \State Update $c^{\mathrm{pos}}_\ell$, $c^{\mathrm{trk}}_\ell$ from committed frames
      \State Incorporate received user pushes
      \State Estimate object states; extrapolate over chunk $\ell+1$
      \For{each predicted approaching contact $(i,j)$, in temporal order}
        \State Query the VLM for $\hat m,\hat e$ of new objects (cached)
        \State Compute $\Delta\mathbf v_i,\Delta\mathbf v_j$ by \cref{eq:supp:impulse}
        \State Update extrapolated velocities; re-predict later contacts
      \EndFor
      \State Paint contact increments into $c^{\mathrm p}_\ell$
      \State Generate and decode chunk $\ell+1$ conditioned on $c_\ell$
    \EndFor
  \end{algorithmic}
\end{algorithm}

\paragraph{Contact prediction.}
From the depth and tracks of committed chunks, the controller fits each
object's camera-space centroid $\mathbf p_k$, velocity $\mathbf v_k$, and
acceleration over the last chunk, adds any user increment already received
for the next chunk, and extrapolates over that chunk, translating each
object's last mask by its projected displacement. A contact between
objects $i$ and $j$ is predicted at the first frame $t_c$ at which their
translated silhouettes overlap and their depth intervals, estimated from
the visible depth and each object's metric extent, overlap within a margin
$\epsilon_z$. The contact normal follows the line between centroids,
$\mathbf n=(\mathbf p_i-\mathbf p_j)/\|\mathbf p_i-\mathbf p_j\|$, pointing
from $j$ to $i$; this is exact for spheres and approximate for other
shapes. The closing speed $v_n=(\mathbf v_j-\mathbf v_i)\cdot\mathbf n$ is
positive when the objects approach each other. Only approaching contacts
($v_n>0$) trigger impulses, and a pair that has received an impulse is
excluded until its masks separate, which prevents repeated impulses while
the objects remain in contact. Resting contacts are left to the
generator's learned behavior.

\paragraph{Object properties.}
The first time an object enters a contact, the VLM
(Qwen2.5-VL-3B~\cite{qwen25vl}) receives a crop of the object and its
approximate metric extent from the depth map, and returns a mass estimate
$\hat m_k$ and a restitution class mapped to
$\hat e_k\in\{0.8,0.5,0.2\}$. Answers are cached per object for later
interactions.

\paragraph{Contact response.}
With combined restitution $\hat e=\hat e_i\hat e_j$, as in common
rigid-body engines~\cite{pybullet}, a frictionless central collision gives
\begin{equation}
\begin{aligned}
\Delta\mathbf v_i&=\phantom{-}(1+\hat e)\tfrac{\hat m_j}{\hat m_i+\hat m_j}\,v_n\mathbf n,\\
\Delta\mathbf v_j&=-(1+\hat e)\tfrac{\hat m_i}{\hat m_i+\hat m_j}\,v_n\mathbf n,
\end{aligned}
\label{eq:supp:impulse}
\end{equation}
clipped to $[-V_{\max},V_{\max}]^3$. Before clipping, the objects separate
along $\mathbf n$ with relative normal speed $\hat e\,v_n$ and total
momentum is conserved. Since increments take effect one chunk after they
are painted, a contact predicted at frame $t_c$ of chunk $\ell+1$ is
written into the map of chunk $\ell$ on both objects' masks at frame
$t_c-4$; contacts must therefore be \emph{predicted} rather than detected.
Contacts within a chunk are resolved in temporal order, and after each
one the updated velocities are used to re-predict later contacts, so
collision chains can propagate within a chunk. Simultaneous contacts on
the same object have their increments summed, approximating coupled
multi-body resolution.

\section{Additional Qualitative Results}
\label{sec:supp:qualitative}

\begin{figure*}[p]
  \centering
  \includegraphics[width=\textwidth,height=0.84\textheight,keepaspectratio]{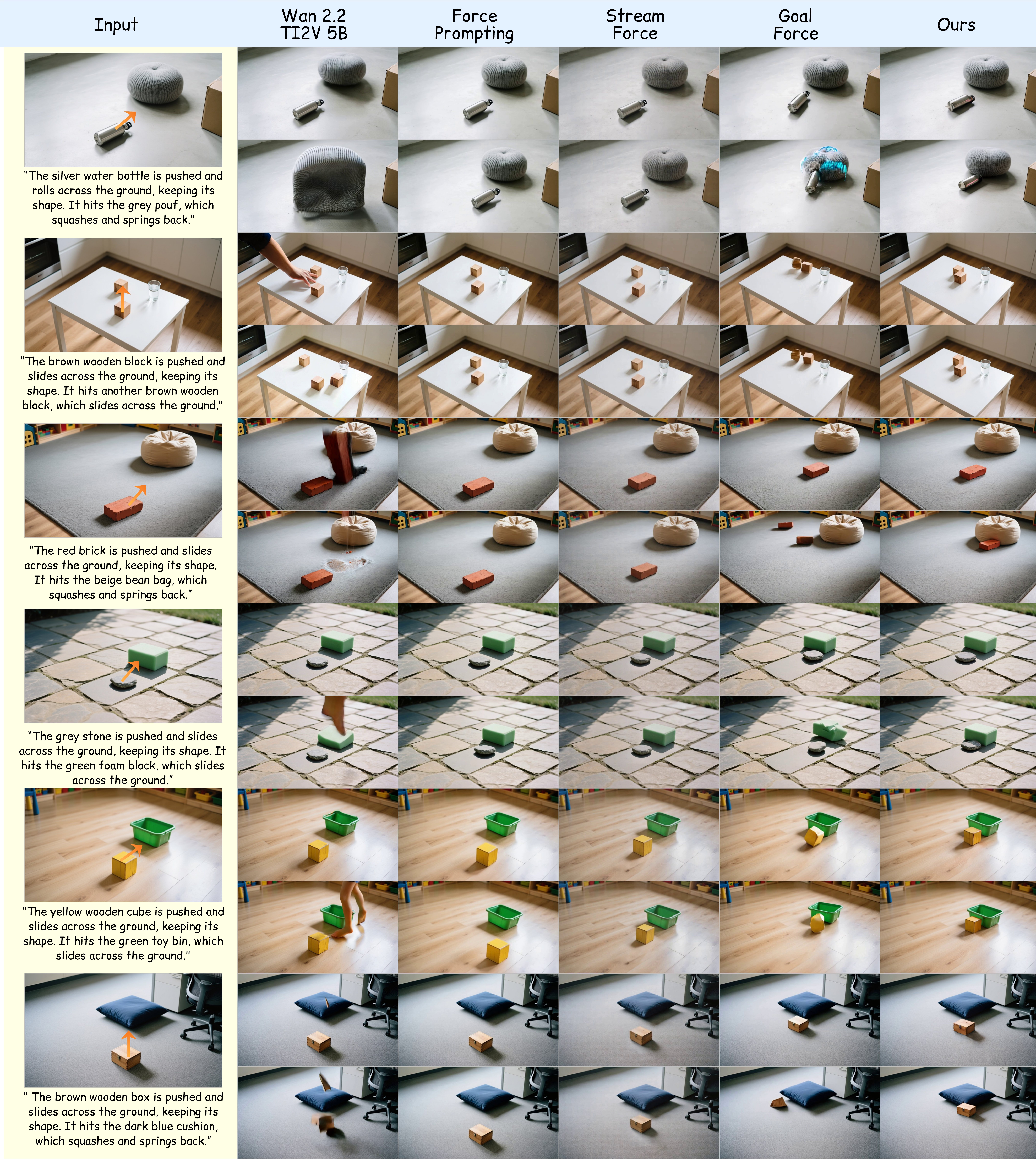}
  \caption{\textbf{Additional comparisons on push-driven interactions.}
  The input column shows the first frame, the push (arrow), and the prompt. For each method, the upper and lower rows show the middle and final frames.}
  \label{fig:supp:interactions}
\end{figure*}

\begin{figure*}[p]
  \centering
  \includegraphics[width=\textwidth,height=0.86\textheight,keepaspectratio]{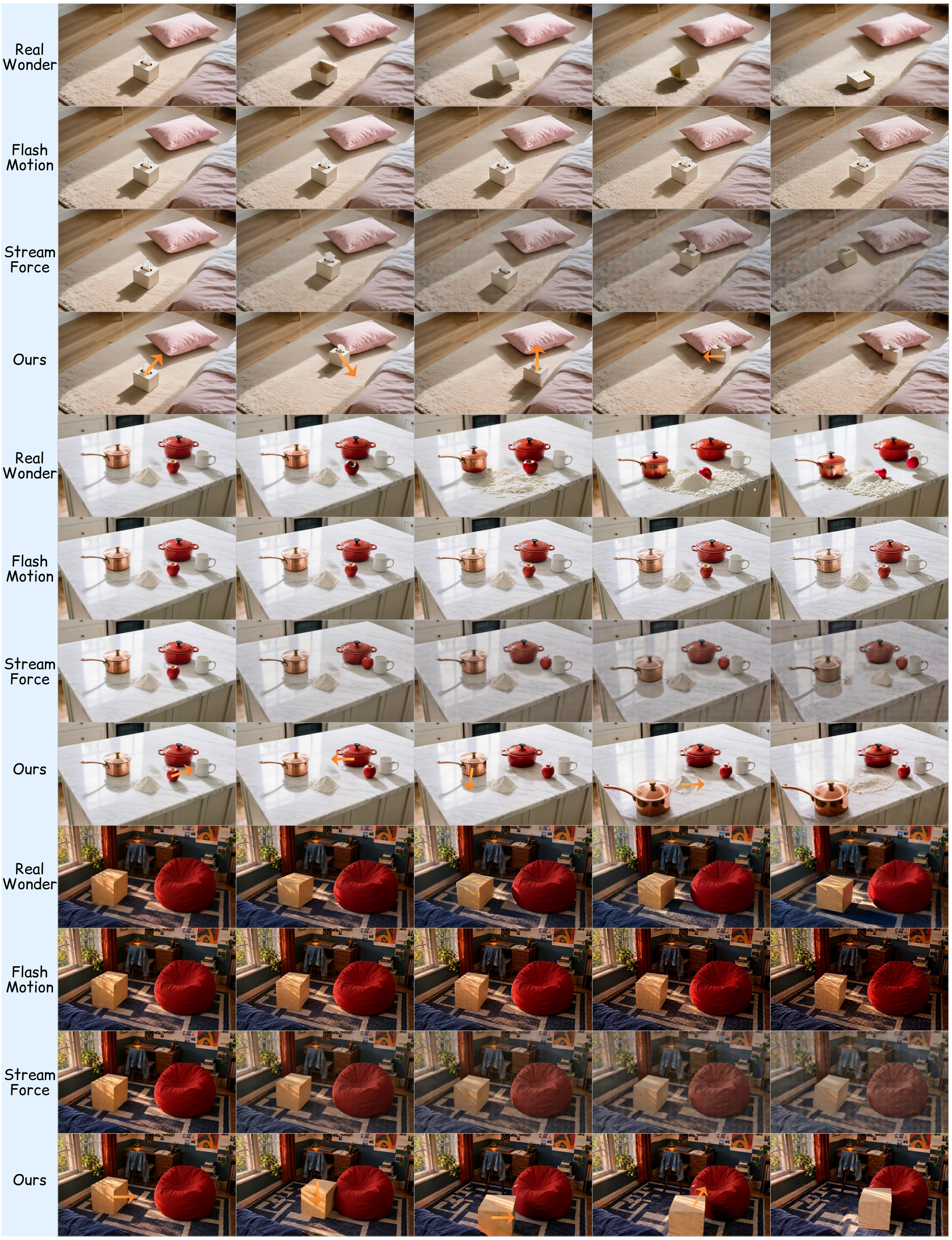}
  \caption{\textbf{Additional streaming comparisons (I).} Each block of four
  rows is one rollout per method, with frames advancing from left to right.
  Arrows, drawn on our rows, mark the pushes issued to all methods during
  generation.}
  \label{fig:supp:stream_a}
\end{figure*}

\begin{figure*}[p]
  \centering
  \includegraphics[width=\textwidth,height=0.86\textheight,keepaspectratio]{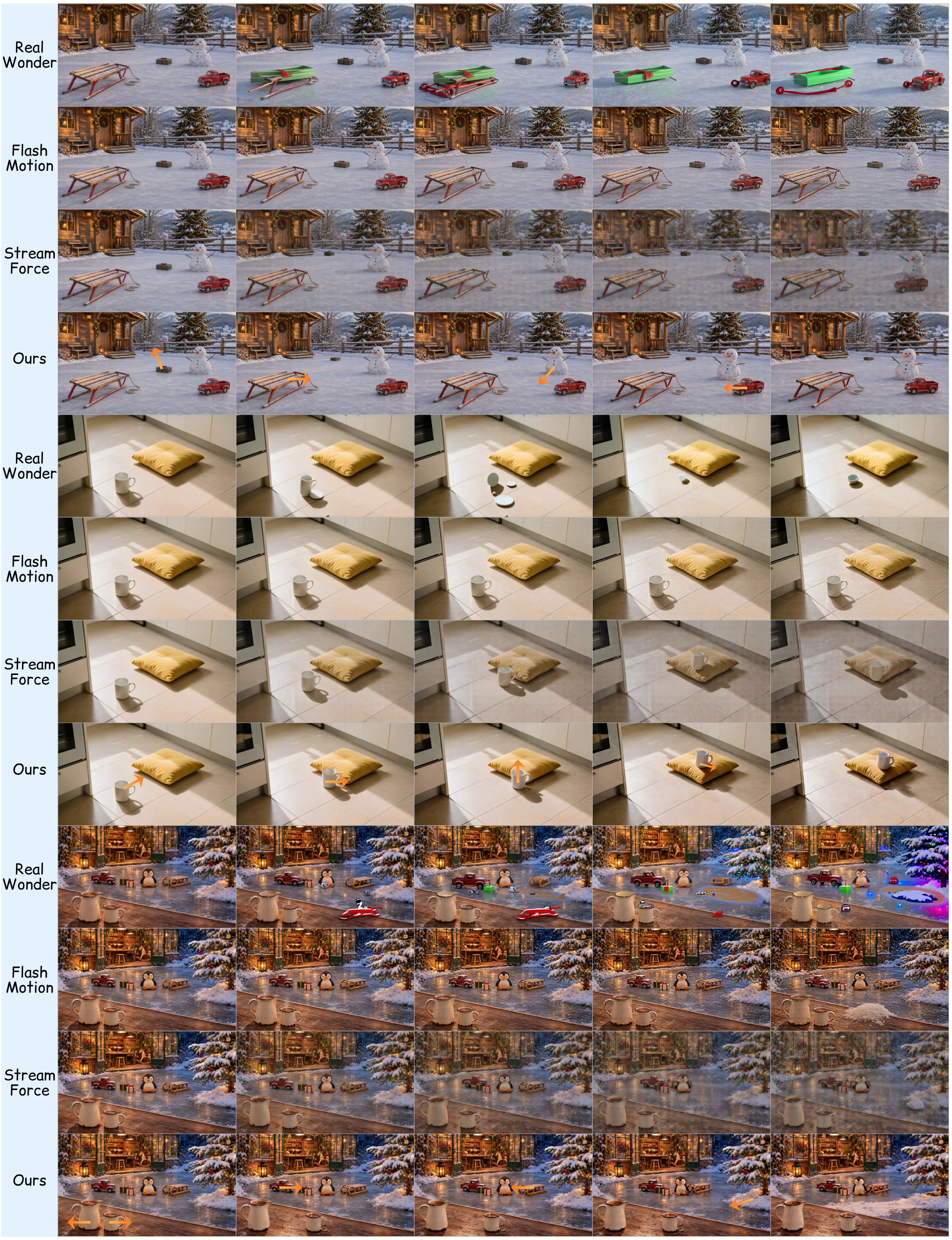}
  \caption{\textbf{Additional streaming comparisons (II).} Layout as in
  \cref{fig:supp:stream_a}.}
  \label{fig:supp:stream_b}
\end{figure*}

This section extends the qualitative studies of the main paper:
\cref{fig:supp:interactions} complements Fig.~4,
\cref{fig:supp:stream_a,fig:supp:stream_b} complement Fig.~5, and
\cref{fig:supp:awareness} adds baselines to Fig.~6. All methods, namely
Wan2.2~\cite{wan22}, Force Prompting~\cite{gillman2025forceprompting},
Goal Force~\cite{goalforce}, StreamForce~\cite{wang2026streamforce}, and,
for streaming, RealWonder~\cite{liu2026realwonder} and
FlashMotion~\cite{li2026flashmotion}, receive the same image, prompt, and
pushes.

\paragraph{Push-driven interactions.}
\Cref{fig:supp:interactions} shows six scenes in which a pushed object
strikes a second object, in most cases of a different material: a steel
bottle and a knitted pouf, two wooden blocks, a brick and a bean bag, a
stone and a foam block, a wooden cube and a plastic toy bin, and a wooden
box and a cushion.
The failure modes identified in the main paper recur across all six scenes.
Wan2.2 adds content instead of executing the push: a hand, a foot, or a pair
of legs enters the frame in three scenes, the pouf swells until it fills the
view, and the wooden box tumbles and breaks apart. Force Prompting leaves the
pushed object nearly in place, and StreamForce moves it only slightly, so in
both cases the contact described by the prompt never occurs. Goal Force
produces motion but violates rigid-body behavior: the brick splits in two,
the foam block and the wooden cube deform, the wooden box collapses into a
wedge, and the pouf dissolves into colored artifacts. Our model moves each
pushed object along the push until it reaches the target and keeps both
objects intact through the contact.

\paragraph{Streaming interactions.}
\Cref{fig:supp:stream_a,fig:supp:stream_b} show six streaming rollouts with
four to five pushes each. Pushes are issued while generation is in progress,
often to an object that has already moved, and a single rollout can involve
several objects of different materials, \eg, an apple, a cast-iron pot, a
copper saucepan, and a heap of flour on the same counter. The three
baselines fail in distinct ways. RealWonder simulates reconstructed 3D
geometry, so reconstruction errors become visible as soon as an object
moves: the tissue box folds into an open tray, the mug shatters into flat
discs and disappears, the sled turns into a green boat, the apple splits,
and the winter table fills with objects that were never in the input, such as a toy plane and glowing rings. FlashMotion keeps every scene essentially static. StreamForce moves some objects but accumulates error over the rollout, with haze, loss of contrast and texture, and ghosted objects by the
final frames. Our model executes each push at the object's current position,
produces both rigid sliding and the spreading of granular flour and snow,
and preserves the appearance of the scene for the full rollout.

\paragraph{Physics awareness.}
\Cref{fig:supp:awareness} varies one factor at a time. Force Prompting and
StreamForce largely ignore it: the rock travels the same short distance at
every intensity, the cue ball and the shaped objects barely move, and
StreamForce moves all three boxes equally far. Goal Force reacts but not in
proportion, carrying the rock to the cushion even at $1\times$, sending the
cue ball past or short of the target balls, toppling the pouf, and driving
every box into the large box regardless of material. Only our model yields
graded outcomes: travel grows with intensity, the cue ball reaches the red,
yellow, and blue balls in turn, the basketball rebounds while the can
arrives slowly and the container stops short, and the steel, leather, and
cardboard boxes travel in order of their estimated masses.

\clearpage
\twocolumn[{%
  \centering
  \includegraphics[width=\textwidth]{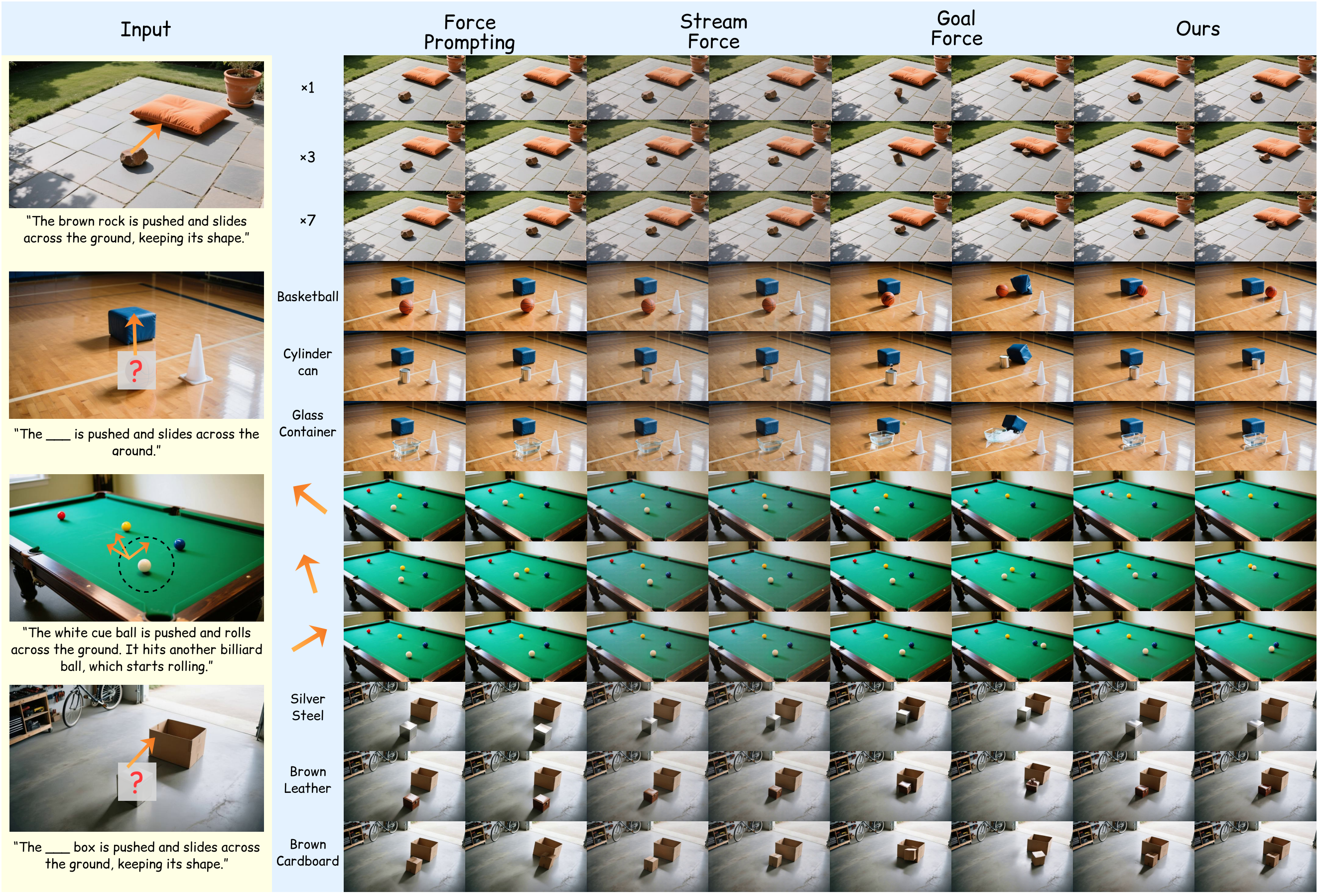}
  \captionsetup{hypcap=false}
  \captionof{figure}{\textbf{Physics awareness compared with baselines.}
  Each row changes a single factor of the input on the left, and each
  method shows the middle and final frames. From top to bottom: push
  intensity ($1\times$, $3\times$, $7\times$), object shape
  (basketball, can, glass container) under the same push, push direction, and material (steel, leather, cardboard). A question mark in the input marks the object that is swapped across rows.}
  \label{fig:supp:awareness}
  \vspace{1.5em}
}]

\end{document}